\pdfoutput=1

\documentclass[11pt]{article}

\usepackage[]{EMNLP2023}

\usepackage{times}
\usepackage{latexsym}

\usepackage{amsmath}
\usepackage{amssymb}
\usepackage{bbm}
\usepackage{booktabs}
\usepackage{multirow}
\usepackage{graphicx}
\usepackage{rotating}
\usepackage{tabularx}
\usepackage[T1]{fontenc}

\usepackage[utf8]{inputenc}

\usepackage{microtype}

\def\MODEL{\textsc{HistAlign}}
\def\TRIME{\textsc{Trime}}

\title{\MODEL{}: Improving Context Dependency in Language Generation \\ by Aligning with History}

\author{David Wan $\;\;\;\;\;\;$ Shiyue Zhang   $\;\;\;\;\;\;$ Mohit Bansal \\
  UNC Chapel Hill  \\
  {\tt \{davidwan, shiyue, mbansal\}@cs.unc.edu} 
}

\begin{document}
\maketitle
\begin{abstract}
Language models (LMs) can generate hallucinations and incoherent outputs, which highlights their weak \emph{context dependency}. Cache-LMs, which augment LMs with a memory of recent history, can increase context dependency and have shown remarkable performance in diverse language generation tasks. However, we find that even with training, the performance gain stemming from the cache component of current cache-LMs is suboptimal due to the misalignment between the current hidden states and those stored in the memory. In this work, we present \MODEL{}, a new training approach to ensure good cache alignment such that the model receives useful signals from the history. We first prove our concept on a simple and synthetic task where the memory is essential for correct predictions, and we show that the cache component of \MODEL{} is better aligned and improves overall performance. Next, we evaluate \MODEL{} on diverse downstream language generation tasks, including prompt continuation, abstractive summarization, and data-to-text. We demonstrate that \MODEL{} improves text coherence and faithfulness in open-ended and conditional generation settings, respectively. \MODEL{} is also generalizable across different model families, showcasing its strength in improving context dependency of LMs in diverse scenarios.\footnote{
Our code is publicly available at \url{https://github.com/meetdavidwan/histalign}}
\end{abstract}

\begin{figure}[!t]
    \centering
    \includegraphics[width=\columnwidth]{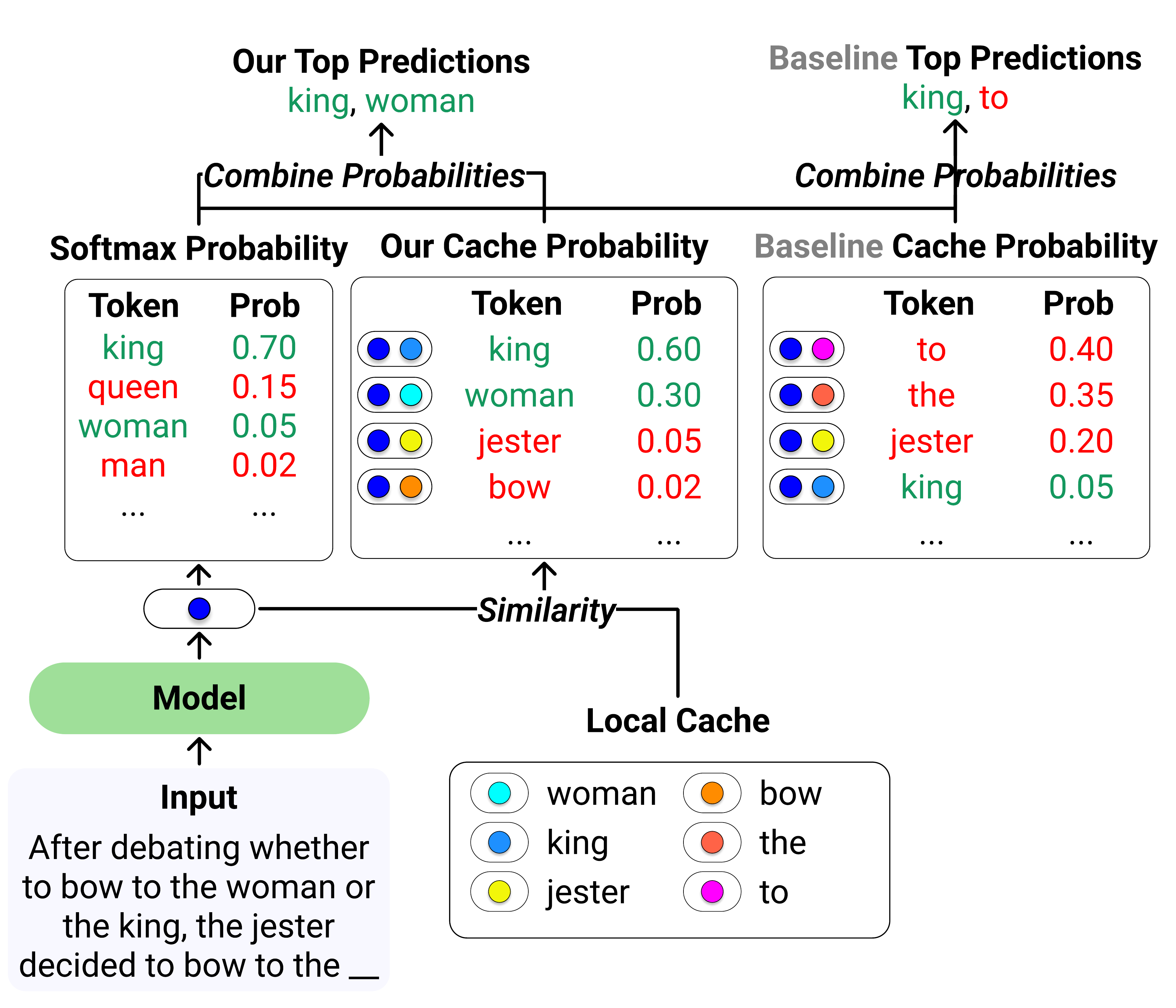}
    \vspace{-5pt}
    \caption{An illustration of \MODEL{} and baseline cache-LM. The input example is from~\citet{chang-mccallum-2022-softmax}. Our \MODEL{} is able to assign high probabilities to both \textit{king} and \textit{woman}, and thus is able to tune down the weight of the hallucinated token \textit{queen} from the softmax probability. Current cache language models (baseline) give high probabilities to irrelevant tokens in the cache and thus are at risk of producing hallucinated or incoherent tokens.}
    \label{fig:figure}
\end{figure}

\section{Introduction}
Language modeling (LM), or language generation, requires decent context dependency. For both open-ended and conditional generation tasks, we want the model generation to be consistent with its previous generation or the input context. However, incoherence and hallucination problems are pervasive in current model generations~\cite{Holtzman2020The, cao2018faithful, maynez-etal-2020-faithfulness}, which suggests the weak context dependency of LMs.

Cache language model~\cite[Cache-LM]{grave-2017-improving} is a simple yet effective method to improve context dependency by equipping LM with an additional memory of recent history (local context) and enabling it to directly ``copy'' from the history. Such models showed considerable improvement in language modeling and downstream generation tasks \cite{ merity2017pointer, see2017get}. However, since the introduction of Transformers~\cite{transformer}, local memory has been less used due to the powerful self-attention mechanism, and more works have been focusing on leveraging long-term or external memory~\cite{Khandelwal2020Generalization, yogatama-etal-2021-adaptive}. Nonetheless, \citet{zhong-etal-2022-training} showed that using local memory on top of a Transformer is still beneficial.

In this paper, we focus on applying local cache to Transformer-based LMs and show that better alignment of the cache component leads to stronger gains. First, we show that cache-LM theoretically breaks the softmax bottleneck~\cite{yang2018breaking} that limits the capacity of any parametric LM to model highly context-dependent natural language. Then, we find that, in current cache-LMs, the signals provided by the memory component are minor, even when using the cache component during training~\cite{zhong-etal-2022-training}. We hypothesize that the main bottleneck comes from the misalignment of the current hidden states and those in the memory, because of which more relevant memories are not given higher weights than less relevant ones. We demonstrate this problem through a synthetic task: Ambiguous Template \cite{chang-mccallum-2022-softmax}, an example of which is shown in \autoref{fig:figure}. When asking the model to predict the next word given the context ``\textit{After debating whether to bow to the woman or the king, the jester decided to bow to the \_\_} ,'' current cache-LM does not give the highest probabilities to the desired words \textit{king} and \textit{woman}. Instead, we find that irrelevant words, such as \textit{to} and \textit{jester} have high cache probabilities. When combining these probabilities with the original softmax, the desired words cannot be ranked as top tokens. We find that this problem exists in pre-trained LMs of various sizes, fine-tuned models, as well as models with cache augmented.

Next, we address this misalignment issue by proposing a new fine-tuning scheme, \MODEL{}, in which we augment the LM training objective with a contrastive loss to encourage the model to align the current hidden states with those in the history. As shown in \autoref{fig:figure}, our cache component gives higher probabilities for \textit{king} and \textit{woman} than other less relevant words in the cache. 
Unlike the typical contrastive loss that treats all negative examples equally, we propose to learn a ranking of negative tokens, i.e., more semantically similar tokens are ranked higher. As shown in~\autoref{fig:train_example}, when we align the space for the token \textit{housing}, we want words such as \textit{accommodations} to be closer than less relevant words like \textit{children}. Hence, the cache can also be useful even when the exact target word is not present in the history. We demonstrate the stronger cache performance of \MODEL{} through the synthetic ambiguous template task and showcase its strength in improving coherence for open-ended prompt continuation and faithfulness for abstractive summarization and data-to-text. 

To summarize, our contributions are as follows:
\begin{itemize}
    \item We discuss why cache-LM with local memory can improve context dependency through a softmax bottleneck lens.
    \item We show the misalignment problem present in current cache language models and their training strategy.
    \item We propose a new training method, \MODEL{}, based on order-informed contrastive learning, 
    which alleviates the misalignment problem and makes better use of memories.
    \item We demonstrate that \MODEL{} improves the coherence of open-ended generation as well as the faithfulness of conditional generation, and it works across different model families and adds little computational overhead.
\end{itemize}

\section{Related Work}
\paragraph{Cache-LM and Pointer Network.} Adding a cache component to a language model (LM) was first introduced for speech recognition~\cite{kuhnvv1990cache}. \citet{grave2017improving} extended this idea to RNN-based neural LM, which they call \emph{neural cache-LM}. Cache-LM predicts the next token by combining the RNN model's outputs with the similarities between the cache and the current hidden state. The cache saves tuples of hidden state and next token prediction, i.e., ($h_i$, $x_{i+1}$), from recent history (see Section~\ref{sec:cache_lm}). Essentially, the cache component enables the model to copy tokens from the history. Similar to cache-LM, a pointer network~\cite{vinyals2015pointer, merity2017pointer} also combines generating and copying of tokens but uses $h_i$ as a representation of $x_i$ (instead of $x_{i+1}$). This means that a pointer network requires learning additional transformations between the current representation and those in the past and a gating component for interpolation \cite{merity2017pointer, see2017get}.\footnote{Depending on the implementation, the model can have additional numbers of parameters that are quadratic to the number of hidden size for the projection matrix (for example, \citet{see2017get} uses the concatenation of four hidden states for the gating module).}
In contrast, cache-LM doesn't need extra parameters to be learned and can be applied directly at testing time. It is more efficient to be used for larger cache sizes (i.e., extending cache-LM to long-term and external memory), and has been shown to perform better than pointer-network \cite{grave-2017-improving, zhong-etal-2022-training}.

While cache-LM can be directly applied at test time, a recent work~\cite{zhong-etal-2022-training} showed that it leads to more improvement when using cache during training time as well. Nonetheless, such proposed learning objectives for cache-LMs usually only provide distant supervision to the cache component. In contrast, we introduce direct supervision to the cache, which aligns the current representation with its history.  

\paragraph{LM with Local or External Memory.} Cache-LM and pointer network were originally proposed to only use hidden states from the local context, i.e., previous tokens in the input context. Though this technique has been proven to be helpful for language modeling and other language generation tasks~\cite{gulcehre-etal-2016-pointing, grave2017improving, merity2017pointer, see2017get}, it has been less used after the Transformer architecture became popular, because the self-attention mechanism can attend to any token in the input context. Therefore, many works~\cite{grave2017unbounded, Khandelwal2020Generalization, yogatama-etal-2021-adaptive, zhong-etal-2022-training, min2022nonparametric} proposed to use long-term or external memory beyond local context by applying retrieval techniques. Though our work can be extended to the external cache setting, we focus only on incorporating local memory, and we show that local memory is still helpful on top of Transformer because it breaks the softmax bottleneck~\cite{yang2018breaking} of parametric language models. A concurrent work~\cite{chang2023revisiting} also demonstrates how a pointer network breaks softmax bottleneck by examples and empirical results, while we discuss this in a more mathematical way in Section~\ref{sec:break_softmax}.

\paragraph{Context Dependency in Language Generation.} Existing language generation models demonstrate weak context dependency. For open-ended generation tasks, \citet{Holtzman2020The} pointed out that strong LMs can produce very incoherent text following an input prompt. This incoherence issue has also been long observed in the story generation literature~\cite{rashkin-etal-2020-plotmachines, alabdulkarim-etal-2021-automatic}. For conditional generation tasks, for example, summarization, \citet{cao2018faithful, maynez-etal-2020-faithfulness} showed that around 30\% and 70\% model-generated summaries contain hallucinations for two popularly used summarization datasets, respectively. Similar unfaithfulness problems have also been seen in data-to-text generation~\cite{chen-etal-2020-logical}, machine translation~\cite{weng-etal-2020-towards}, etc. Though many approaches have been introduced to alleviate incoherence~\cite{contrastive_decoding} or unfaithfulness~\cite{cao-wang-2021-cliff, wan-bansal-2022-factpegasus}, in this work, we explore a simple yet general cache-LM method to increase context dependency for diverse tasks. The concurrent work~\cite{chang2023revisiting} uses pointer network type of architectures to improve next-word distribution and summarization factuality. They modify the softmax head by using additional context-dependent embeddings. In contrast, we simply apply the original cache-LM architecture and improve it with a novel training objective.

\section{Preliminaries}

\subsection{Language Modeling}
\label{sec:lm}
We focus on autoregressive language modeling (LM). Here, for simplicity, we assume that the LM is decoder-only, i.e., the context of the current step is the generated tokens of previous steps. We show that the same approach can easily be generalized to encoder-decoder models in Section~\ref{sec:encoder-decoder}. Given the context $c_t = x_1,...,x_{t-1}$, the probability of next token $x_t=w$  is predicted by a softmax head:
\begin{equation}
P_{lm}(w|c_t) \propto \text{exp}(h_t^\top e_w)
\end{equation}
where $e_w$ is the output embedding of token $w$ and $h_t$ is the output context vector (hidden state) from the model at the $t$-th step. The model is trained by minimizing the cross-entropy loss: $l_{xe} = -\sum_t \log P_{lm}(x_t|c_t)$.

\subsection{Cache Language Models}
\label{sec:cache_lm}

Cache language models augment a memory component to language models. Following \citet{grave2017improving}, we consider cache to be a list of tuples of context vector and target token, $(h_i,x_i)$.  Assume we only consider the history of the local context, then the local memory of $t$-th step is written as:
\begin{equation}
\mathcal{M}_{\text{local}} = \{( h_i, x_i )\}_{1 \leq i \leq t-1}
\end{equation}
Then, the next-token prediction aggregates the logits from the softmax head and the similarities between $h_t$ and those saved in the memory:
\begin{equation}\label{eq:cache_prob}
\begin{aligned}
&P_{clm}(w|c_t) \propto  \text{exp}(h_t^\top e_w) + \\
&\sum_{ (h_i,x_i) \in \mathcal{M}_{\text{local}}} \mathbbm{1}_{\{ x_i=w \}} \text{exp}( \text{sim}(h_t, h_i ))
\end{aligned}
\end{equation}
where $\text{sim}(\cdot,\cdot)$ can be an arbitrary similarity function. Here, we follow \citet{zhong-etal-2022-training} and use the scaled dot product: $\text{sim}(h_1,h_2) = \frac{h_1 \cdot h_2}{\sqrt{d}}$, where $d$ is the hidden dimension size.

While \citet{grave2017improving} only incorporated cache during evaluation, \TRIME{}~\cite{zhong-etal-2022-training} showed that it brings more benefits when also incorporated during training, i.e., minimizing $l_{trime} = -\sum_t \log P_{clm}(x_t|c_t)$. Here, we also use cache in both training and evaluation, but we improve the training objective by introducing direct supervision on the cache (see  Section~\ref{sec:histalign}).

\begin{figure*}[!t]
    \centering
    \includegraphics[width=0.95\textwidth]{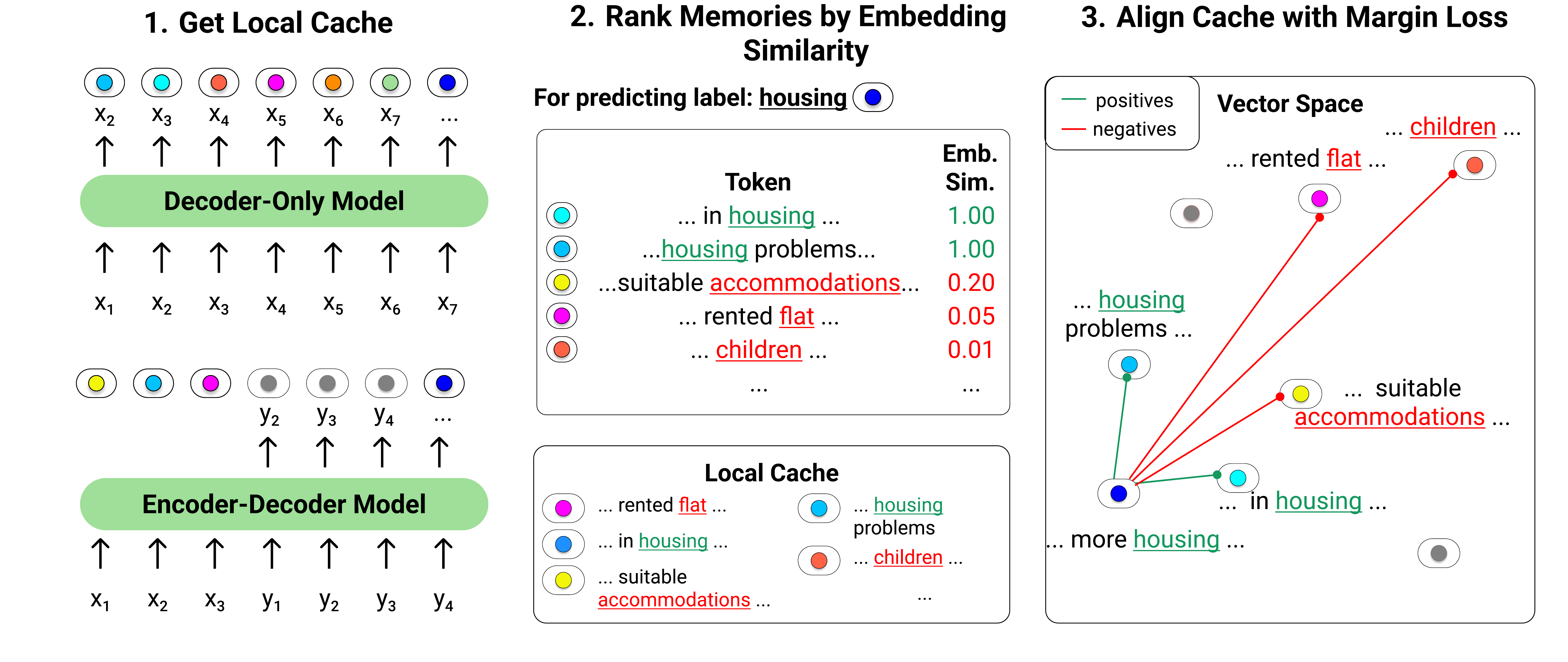}
    \vspace{-5pt}
    \caption{Illustration of our \MODEL{} training approach. We first get local cache by combining the hidden states in local context with their target tokens, and then rank them according to embedding similarity. The ranked memories are then used to train with the margin loss. This ensures that negative yet similar words (e.g. accommodations) will be closer in the vector space than irrelevant words (e.g. children).
    }
    \label{fig:train_example}
\end{figure*}

\section{Our Methodology}

\subsection{Breaking Softmax Bottleneck}
\label{sec:break_softmax}
We first want to connect using local memory with the softmax bottleneck problem~\cite{yang2018breaking} and show that Transformer's self-attention cannot break this bottleneck, while the local cache can.

Parametric autoregressive language models (Section~\ref{sec:lm}), including Transformer-based LMs, use a softmax function operating on context vectors (or hidden states) $\mathbf{H} \in \mathbb{R}^{N \times d}$ and output embedding matrix $\mathbf{E} \in \mathbb{R}^{V \times d}$. $N$ is the number of contexts, assuming every token in the training set has a different context, then $N$ is the number of tokens in the training set. $V$ is the vocabulary size, and $d$ is the hidden dimension size. Then, the next token probabilities form a log-probability matrix $\mathbf{A} \in \mathbb{R}^{N \times V}$ ($A_{tw}=\log P(w|h_t)$). Ideally, since every context is unique, the rank of $\mathbf{A}$ should be as large as $V$ (assuming $V<N$). However, as $\mathbf{A}$ is roughly equivalent to $\mathbf{H}\mathbf{E}^\top$, its rank is strictly upper bounded by hidden size $d$ (please refer to \citet{yang2018breaking} for the formal proof). This low-rank problem greatly limits the LM's capacity to model highly context-dependent natural language. This can be seen in \autoref{fig:figure}, where \textit{queen} achieves higher probability than \textit{woman}. The reason for LM's difficulty in such bimodal distribution, as explained in \citet{chang-mccallum-2022-softmax}, is that the four words \textit{king}, \textit{woman}, \textit{man}, \textit{queen} tend to form a parallelogram in the embedding space, and if the model's hidden state wishes to be close to the output embeddings of \textit{king} and \textit{woman}, it will also be close to those of \textit{man} and \textit{queen}.

To break this bottleneck, one simple solution is to increase $d$, as we see larger models usually have better performance. Another solution proposed by \citet{yang2018breaking} and extended by~\citet{Sigsoftmax, Mixtape, chang-mccallum-2022-softmax} is to use multiple softmax heads -- mixture of softmax (MoS), e.g., $P(w|h_t) \propto \exp(h_t^{(1)\top} e_w) + \exp(h_t^{(2)\top} e_w) + \exp(h_t^{(3)\top} e_w)$. Each $h_t^{(k)}$ is a different context vector. However, adding softmax heads is fairly computationally expensive. Comparing MoS to Eq.~\ref{eq:cache_prob}, we can see that adding $\text{exp}(h_t^\top e_w)$ and $\text{exp}( \text{sim}(h_t, h_i ))$ resembles MoS without adding extra softmax heads. Another way to understand this connection is that when using local memory, $\mathbf{A}$ is roughly equivalent to $\mathbf{H}\mathbf{E}^\top + \mathbf{H}\mathbf{H}_c^\top$, where $\mathbf{H}_c$ are the hidden states in the local context.\footnote{This is because the logits under the cache-LM setting are the sum of the original token logits $HE^\top$ (first term of Eq.~\ref{eq:cache_prob}) and the cache logits $HH_C^\top$ (the second term of Eq.~\ref{eq:cache_prob}), and property 2 of \citet{yang2018breaking} showed that the log-probability matrix and logit matrix have similar ranks.} Assuming $\mathbf{E}_c=\mathbf{E}+\mathbf{H}_c$, $\mathbf{A}$ becomes $\mathbf{H}\mathbf{E}_c$. Different from $\mathbf{E}$, $\mathbf{E}_c$ is no longer a static output embedding matrix of size $V \times d$ but a context-dependent embedding tensor of size $N \times V \times d$. Hence, the rank of $\mathbf{A}$ is no longer upper bounded by $d$. Note that this connection also holds for using long-term or external memories.

\subsection{\MODEL{}}\label{sec:histalign}
Cache-LM combines the original softmax probabilities with the cache probabilities by aggregating the similarity scores between the current hidden state and those in the cache. To use the cache module effectively, the similarity function $\text{sim}(\cdot, \cdot)$ plays an important role in Eq.~\ref{eq:cache_prob}. If the similarities between the current hidden state and less relevant memories are higher than more relevant ones, it would steer the model away from selecting the most useful information from the cache. By assigning a high probability to the correct local memories, e.g., those corresponding to \textit{king} and \textit{woman} in the example of \autoref{fig:figure}, we can ensure that when the probabilities are combined, they will be scored higher than irrelevant and hallucinated tokens.
However, we find that even when directly maximizing $\log P_{clm}$~\cite{zhong-etal-2022-training}, there is no guarantee that the current representations are well aligned with relevant information stored in the memory, as shown by the baseline probabilities in \autoref{fig:figure} (see Section~\ref{sec:result_ambiguous} for more details). 

Hence, to deal with this misalignment, we propose a new contrastive objective that encourages higher similarities between the hidden states of similar target tokens. During training, given the current hidden state $h_t$ and the corresponding next token $x_t$, we construct a positive set $\mathcal{P}_t$ from caches by selecting memories with the same target token: 
\begin{equation}
\mathcal{P}_t = \{( h_i, x_i )\}_{x_i=x_t, 1 \leq i \leq t-1}
\end{equation}
All other memories are taken as negative examples. An example is shown in step 2 of \autoref{fig:train_example}. For predicting the token \textit{housing}, we have two previous mentions of the word housing, and the other words, including \textit{flat, children, accommodations, etc.}, are considered as negative. 

In the typical contrastive loss, such as InfoNCE \cite{oord2019representation}, all negative examples are treated equally. However, we hope to learn an ordering of the negative examples --  more similar examples are ranked higher than less similar ones. In the example in \autoref{fig:train_example}, \textit{accommodations} is more similar to \textit{housing} than \textit{children}. This ensures that even when predicting words that do not have previous mentions in the local cache, our model can still output a reasonable alternative.

To achieve this, we construct a ranking of memories by computing the cosine similarities between the embedding of the current target word and the embeddings of words in the cache, i.e., $\text{cosim}(e_t,e_i)$. After sorting tokens from the most similar w.r.t. semantic similarity to the least, we use the following max-margin loss~\cite{liu-etal-2022-brio}:
\begin{equation}
\begin{aligned}
l_{cont.} = \sum_t \sum_{i \in \mathcal{P}_t} \sum_{j>i, j \notin \mathcal{P}_t} & \max{(}0,  \text{sim}(h_t,h_{j}) \\
 & - \text{sim}(h_t, h_{i}) + \lambda_{i,j} )
\end{aligned}
\end{equation}
where $\lambda_{i,j} = (j-i) \lambda$, and $\lambda$ is the margin tuned based on validation loss.

The final objective of \MODEL{} is a combination of the original LM cross-entropy loss $l_{xe}$ and this ranking-based contrastive loss:
\begin{equation}
    l_{histalign} = l_{xe} + \alpha l_{cont.}
\end{equation}
where $\alpha$ is a tunable weight of the contrastive loss.
Note that during the inference time, we use Eq.~\ref{eq:cache_prob}.

\subsection{Extension to Encoder-Decoder Models}\label{sec:encoder-decoder}
\MODEL{} can be easily adapted to encoder-decoder models. For conditional generation tasks, the target text is usually short, hence, coherence is not a big issue. What is more crucial is whether the target generation stays true to the input context, e.g., the input document for summarization or the input table for data-to-text. Therefore, we define the local cache to be the input tokens and their corresponding encoder hidden states, as opposed to the output tokens and decoder hidden states for decoder-only models. We then calculate the similarity between the current decoder hidden state with those encoder hidden states stored in the cache.

\section{Experimental Setup}
Here, we describe the tasks and the experimental setups.
Please refer to Appendix~\ref{sec:experimental_appendix} for more details.

\subsection{Tasks and Datasets}

\textbf{Ambiguous Template} is a useful synthetic dataset collated by \citet{chang-mccallum-2022-softmax}, in which each example is generated using templates with diagonal words\footnote{This refers to words that lie on the diagonal of a parallelogram in the embedding space. For example, for the tuple (queen, king, woman, man), the diagonal words are (king, woman) and (queen, man).} from semantic analogy relations in the Google (English) analogy dataset \cite{NIPS2013_9aa42b31}.
This is a simple yet effective setting to examine whether the model can copy the correct tokens from history and not hallucinate semantically similar tokens, e.g., \emph{queen} and \emph{man} of the example in \autoref{fig:figure}. Since the target words can always be found in the context, we can also evaluate the performance only with the cache component.

\textbf{Open-Ended Generation} evaluates the language modeling capability by asking the model to generate a continuation given a prompt~\cite{Holtzman2020The,su2022a,li2022contrastive}. We use WritingPrompts \cite{fan-etal-2018-hierarchical}, and treat the first 50 tokens as the prompt and allow the model to generate up to 256 tokens using the canonical nucleus sampling ($p=0.95$)~\cite{Holtzman2020The}.

\textbf{Abstractive Summarization} is the task of providing an abridged version of the input document. One crucial problem is `hallucination', where the generated summaries contain facts or entities that are wrong or not present in the document~\cite{cao2018faithful, maynez-etal-2020-faithfulness}. We evaluate on two widely-used English News summarization datasets, XSum \cite{narayan-etal-2018-dont} and CNN/DM \cite{NIPS2015_afdec700}. 

\textbf{Data-to-Text} is the task of describing structured data, where faithfulness is extremely important, as humans do not tolerate any hallucinations in cases such as describing medical reports or financial statistics~\cite{thomson-reiter-2020-gold}. We evaluate on LogicNLG \cite{chen-etal-2020-logical}. 

\subsection{Systems} \label{sec:model}
We use GPT2-small and GPT2-large~\cite{radford2019language} for ambiguous template and prompt continuation, and we use BART-large~\cite{lewis-etal-2020-bart} for both summarization and data-to-text. For all tasks, we choose to finetune pre-trained LMs. The first baseline we compare to is fine-tuning with the original cross-entropy loss ($l_{xe}$ in Section~\ref{sec:lm}), which is named by the original model name in our result tables. Then, we also compare to the most recent cache-LM learning objective, \TRIME{}~\cite{zhong-etal-2022-training} ($l_{trime}$ in Section~\ref{sec:cache_lm}). 

\begin{table*}[!t]
    \centering
    \small
    \begin{tabular}{ l c c c c c c c c |c}
    \toprule
    & \multicolumn{4}{c}{Full} & \multicolumn{4}{c}{Cache-Only} & Full\\
    \cmidrule(lr){2-5} \cmidrule(lr){6-9} \cmidrule(lr){10-10}
    Model & Acc@2 & Acc@5 & Acc@10 & Acc@25 & Acc@2 & Acc@5 & Acc@10 & Acc@25 & Rank \\
    \midrule
    GPT2-Small & 50.00 & 62.20 & 68.80 & $\phantom{0}$76.96 & $\phantom{0}$0.00 &  35.57 & 50.56 & $\phantom{0}$75.71 & $\phantom{0}$762 \\
    \TRIME{} & 46.43 & 56.41 & 76.47 & $\phantom{0}$97.96 & $\phantom{0}$0.00 & 39.89 & 66.51 & \textbf{100.00} & $\phantom{0}$836 \\
    \MODEL{} & \textbf{63.47} & \textbf{72.26} & \textbf{89.71} & \textbf{100.00} & \textbf{58.62}  & \textbf{70.62} & \textbf{79.59} & $\phantom{0}$94.31 & $\phantom{0}$\textbf{854}
     \\
    \midrule
    GPT2-Large & 75.43 & 84.76	& 87.93	& 91.26 & $\phantom{0}$0.05 & 37.83	& 73.88 & \textbf{100.00}  & 1280 \\
    \TRIME{} & 77.40 & 91.10 & 94.56 & 97.17 & $\phantom{0}$0.11	& 22.32	& 84.89	& \textbf{100.00} & \textbf{1377} \\
    \MODEL{} & \textbf{82.22} & \textbf{92.84} & \textbf{96.57} & \textbf{98.34} & \textbf{82.15} & \textbf{92.51}	& \textbf{99.94} & \textbf{100.00} & \textbf{1377} \\
    \bottomrule
    \end{tabular}
    \vspace{-5pt}
    \caption{Results on Ambiguous Template. \MODEL{} achieves the best performance in both full and cache-only settings. We also empirically show that \TRIME{} and \MODEL{} break the softmax bottleneck.}
    \label{tab:ambiguous}
    \vspace{-12pt}
\end{table*}

\begin{table}[!t]
    \centering
    \resizebox{0.95\columnwidth}{!}{
    \begin{tabular}{ l c c c c}
    \toprule
    Model & Acc@2 & Acc@5 & Acc@10 & Acc@25 \\
    \midrule
    LLaMA2-7B & 0 & 0 & 0 & \textbf{100} \\
    \TRIME{} & 0 & 0 & 0 & \textbf{100} \\
    \MODEL{} & \textbf{100} & \textbf{100} & \textbf{100} & \textbf{100} \\
    \bottomrule
    \end{tabular}
    }
    \vspace{-5pt}
    \caption{ Cache-Only results on Ambiguous Template with LLaMA2-7B model. }
    \label{tab:ambiguous_llama}
    \vspace{-12pt}
\end{table}

\subsection{Evaluations}
\textbf{Ambiguous Template.} As a proof-of-concept experiment, we evaluate under both a full setting, using the combined probability in Eq.~\ref{eq:cache_prob}, as well as a cache-only setting, only using the cache similarity scores to predict the next token. We evaluate the performance via the accuracy of having the two diagonal words within the top-$k$ predictions (Acc@$k$), where $k=\{2,5,10,25\}$. Ideally, we want to see 100\% accuracy with $k=2$, which indicates that the two diagonal words are the top 2 choices. Note that when only using the cache, a $k$ value of 50 would achieve perfect accuracy, as it would include the entire local history. In addition, we want to empirically verify that cache LM with local memory can break the softmax bottleneck. To this end, we calculate the rank of log-probability matrix $\mathbf{A} \in \mathbb{R}^{N \times V}$ (Section~\ref{sec:break_softmax}) using 500 examples (concretely, $N=4750$ and $V=50257$ for GPT-2 based models) under the full setting. 

\textbf{Open-Ended Generation.} We mainly evaluate the \emph{coherence} of model-generated continuations. Following~\citet{su2022a}, coherence is approximated by the cosine similarity of the SimCSE~\cite{gao-etal-2021-simcse} sentence embeddings of the prompt and the continuation. In addition, following previous works, we report $n$-gram diversity~\cite{meister+al.pre22} and \textsc{mauve} \cite{NEURIPS2021_260c2432} scores for a more general evaluation. We hope \MODEL{} not to harm diversity and \textsc{mauve}. We also run \textbf{human evaluation} on Amazon MTurk to ask workers to compare the continuations generated by \TRIME{} and \MODEL{}. More details can be found in Appendix~\ref{sec:prompt_human}.

\textbf{Abstractive Summarization.} We mainly evaluate the faithfulness of generated summaries by three widely-used automatic metrics: FactCC \cite{kryscinski-etal-2020-evaluating} and DAE \cite{goyal-durrett-2021-annotating}, which are entailment-based metric; and Entity Precision \cite[P\textsubscript{ENT}]{nan-etal-2021-entity}, which calculates the percentage of entities in the summary that are present in the document. We also report ROUGE-L \cite{lin-2004-rouge} for general content selection evaluation. Similarly, we conduct \textbf{human evaluation}, where we ask crowd workers to judge whether each summary (of 100 randomly selected examples) is faithful and informative. Please refer to Appendix~\ref{sec:summ_human} for more details.

\textbf{Data-to-Text.} We mainly evaluate the faithfulness of model generations by NLI-Acc and SP-Acc \cite{chen-etal-2020-logical} and two more recent metrics -- TAPEX-Acc and TAPAS-Acc \cite{liu-etal-2022-plog}. NLI-Acc is an entailment-based metric pre-trained on TabFact dataset \cite{Chen2020TabFact} using TaBERT \cite{yin-etal-2020-tabert}, and SP-Acc first parses the sentence into a logical program and evaluates the execution accuracy. TAPEX-Acc and TAPAS-Acc are entailment-based metrics trained with TAPEX \cite{liu2022tapex} and TAPAS \cite{eisenschlos-etal-2020-understanding}, respectively. Same as previous works~\cite{chen-etal-2020-logical}, we report BLEU \cite{papineni-etal-2002-bleu} for a surface-level evaluation.

\section{Results}

We verify the strength of \MODEL{} at aligning the cache component and thus improve the next-token prediction on ambiguous template in Section~\ref{sec:result_ambiguous}, coherence in open-ended prompt continuation in Section~\ref{sec:result_prompt}, and faithfulness in abstractive summarization and data-to-text in Section~\ref{sec:result_summ} and Section~\ref{sec:result_data2text}, respectively.

\subsection{Importance of Cache on Ambiguous Template} \label{sec:result_ambiguous}
We show the results of the Ambiguous Template in \autoref{tab:ambiguous}. First, it can be seen that the original GPT2 model has pretty bad performance in the cache-only setting, especially considering Acc@2. This is expected because the original model is fine-tuned using the cross-entropy loss without the cache component involved, and thus applying cache at test time may not be helpful. Second, though \TRIME{}~\cite{zhong-etal-2022-training}
generally outperforms the original model in the full setting, its cache-only Acc@2 and Acc@5 are similar to the original model. Considering that all target words are present in the history, this result indicates that despite the fact that \TRIME{} uses cache during training, its cache component is still misaligned and has limited contributions to the final performance.

In contrast, \MODEL{} achieves high Acc@2 with only the cache module, substantially outperforming the original model and \TRIME{} on both model sizes, which demonstrates the effectiveness of our contrastive loss for aligning memories better. As a result, \MODEL{} outperforms both baselines across all $k$ in the full setting. And the improvement holds for both model sizes, though with smaller gaps for the large model. This observation is consistent with our discussion in Section~\ref{sec:break_softmax} that a larger model with a larger hidden dimension suffers less from the softmax bottleneck, while local memory can help break this bottleneck of any parametric LM. This is also empirically verified by the rank of the log-probability matrix reported in \autoref{tab:ambiguous}, where we see that the rank of the original model is upper-bounded by its hidden dimension (768 for GPT2-small and 1280 for GPT2-large), and having a local cache breaks this bottleneck. Finally, we present two qualitative examples in \autoref{tab:amb_qualitative}. See detailed discussions in Appendix~\ref{sec:results_ambiguous_appendix}.

\paragraph{Experiment on recent LLM.} We also fine-tune LLaMA2 7B model \cite{touvron2023llama}. Interestingly, we find that LLaMA2 achieves 0\% accuracy for Acc@\{2,5,10\} when evaluated zero-shot. After fine-tuning, the model achieves 100\% accuracy without any cache. This is expected, as the task is a simple synthetic task, and the model, compared to GPT2-large, is 10x larger, and the hidden size is 3.2x larger (1280 $\rightarrow$ 4096). Thus, as mentioned in Section~\ref{sec:break_softmax}, the model alleviates the softmax bottleneck due to its larger hidden size.

However, we still observe the two problems with LLaMA2. First, the problem of softmax bottleneck still exists, as the rank of its output log-probability matrix $\mathbf{A}$ is still upper-bounded by its hidden size of 4096, as we find that its empirical rank is 3332. This means that it is still theoretically less expressive than highly context-dependent natural language. Second, \TRIME{} is still not able to make good use of the cache, i.e., misalignment still exists. As shown in the \autoref{tab:ambiguous_llama}, \TRIME{} achieves 0\% accuracy for Acc@\{2,5,10\} under the cache-only setting, which shows that the issue of misalignment is even more apparent for larger language models: Since the token logits perform well enough, the model does not learn to use the cache anymore. Nevertheless, as shown in the table, our training objective can enforce the use of the local cache and achieve 100\% accuracy, which is consistent with our findings from smaller models.

The presence of these two issues showcases that there is still room for improvement on LM’s context dependency, as \MODEL{} outperforms \TRIME{} in making good use of cache.

\begin{table}[!t]
    \centering
    \small
    \begin{tabular}{l c c c}
    \toprule
    Model & diversity & \textsc{mauve} & coherence \\
    \midrule
    GPT2-small & 88.13$_{\pm0.12}$  & 86.62$_{\pm 1.10}$  & 53.77$_{\pm 0.29}$ \\
    \TRIME{} & 88.53$_{\pm 0.14}$ & 86.76$_{\pm 0.58}$ & 57.58$_{\pm 1.05}$ \\
    \MODEL{} & \textbf{90.07}$_{\pm 0.19}$ & \textbf{87.46}$_{\pm 0.80}$ & \textbf{61.30}$_{\pm 0.15}$ \\
    \midrule
    GPT2-large & 88.82$_{\pm 0.07}$ &  86.18$_{\pm 0.94}$ & 52.39$_{\pm 0.10}$ \\
    \TRIME{} & \textbf{90.70}$_{\pm 0.08}$ & \textbf{87.27}$_{\pm 0.85}$ & 53.11$_{\pm 0.19}$ \\
    \MODEL{} & 89.41$_{\pm 0.08}$ & 86.83$_{\pm 1.02}$ & \textbf{53.51}$_{\pm 0.05}$ \\
    \bottomrule
    \end{tabular}
    \vspace{-5pt}
    \caption{Automatic evaluation results of open-ended generation. Numbers are 3-run averages $\pm$ the 95\% confidence intervals.}
    \label{tab:prompt_continuation}
    \vspace{-10pt}
\end{table}

\subsection{Coherence in Open-Ended Generation}\label{sec:result_prompt}

The results of prompt continuation can be found in \autoref{tab:prompt_continuation}. Across both sizes of the model, we observe an improvement in coherence with \TRIME{} and a larger improvement with \MODEL{}. The effect of \MODEL{} is especially prominent for the smaller model, where coherence increases by 7.5 points compared to the original model, and 3.7 points over \TRIME{}. This validates our hypothesis that \MODEL{} can improve the coherence of LMs. When looking at \textsc{mauve}, \MODEL{} improves by 0.8 points and 0.7 points over GPT2 and \TRIME{} respectively when using small models. On the large model, while \TRIME{} achieves the best performance, \MODEL{} still improves over the original model by 0.7 points. A similar trend can be observed for diversity. Holistically, \MODEL{} improves coherence while maintaining similar diversity and \textsc{mauve}.

Besides automatic evaluations, we also conduct a \textbf{human evaluation}, the results of which are shown in \autoref{tab:prompt_continuation_human}. On both fluency and coherence, human raters prefer the continuations by \MODEL{} more than that by \TRIME{}. This confirms the observation from the automatic evaluations that \MODEL{} does improve especially on coherence.

\begin{table}[!t]
    \centering
    \small
    \begin{tabular}{c c c c c c}
    \toprule
    \multicolumn{3}{c}{Fluency} & \multicolumn{3}{c}{Coherence} \\
    \cmidrule(lr){1-3} \cmidrule(lr){4-6}
    Win$\uparrow$ & Tie & Lose$\downarrow$ & Win$\uparrow$ & Tie & Lose$\downarrow$ \\
    \midrule
    46.33 & 36.33 & 17.33 & 48.33 & 32.00 & 19.66 \\
    \bottomrule
    \end{tabular}
    \vspace{-5pt}
    \caption{Human evaluation results of open-ended generation. We conduct a pairwise comparison between \MODEL{} with \TRIME{} (``Win'' means humans prefer our \MODEL{} over \TRIME{}) and show the percentage of passages that are judged as coherent and fluent. \MODEL{} is statistically significantly better ($p<0.05$) than \TRIME{} on fluency and coherence.
    }
    \label{tab:prompt_continuation_human}
    \vspace{-10pt}
\end{table}

\begin{table*}[!t]
    \centering
    \small
    \begin{tabular}{l c c c c | c c c c}
    \toprule
    & \multicolumn{4}{c}{XSum} & \multicolumn{4}{c}{CNN/DM} \\ 
     \cmidrule(lr){2-5} \cmidrule(lr){6-9}
     Model & Rouge-L & FactCC & DAE$\downarrow$ & P\textsubscript{ent} & Rouge-L & FactCC & DAE$\downarrow$ & P\textsubscript{ent}\\
    \midrule
     BART & 36.41 & 22.16 & 67.96 & 72.72 & \textbf{30.63} & 72.63 & 6.98 & 93.53 \\
    \TRIME{} & \textbf{36.50} & 22.94 & 66.34 & 74.25 & 30.60 & 72.65 & 7.08 & 93.39  \\
    \MODEL{} & 35.45 & \textbf{23.07} & \textbf{63.18} & \textbf{75.71} & 29.96 & \textbf{74.93} & \textbf{5.73} & \textbf{93.80} \\
     \bottomrule
    \end{tabular}
    \vspace{-5pt}
    \caption{Performance on abstractive summarization tasks. \MODEL{} consistently improves faithfulness over the two baseline methods on both datasets. }
    \label{tab:summ_results}
    \vspace{-10pt}
\end{table*}

\begin{table}[t]
    \centering
    \small
    \begin{tabular}{l c c c}
    \toprule
     Model & Faithfulness & Informativeness \\ 
    \midrule
    BART & 19.33\phantom{*} & 63.67 \\
    \TRIME{} & 20.00\phantom{*} & 66.33 \\
    \MODEL{} & 26.33* & 65.33 \\
    \bottomrule
    \end{tabular}
    \vspace{-5pt}
    \caption{Human evaluation results on XSum. * indicates that it is statistically significantly better (p < 0.05) than BART. Krippendorff’s $\alpha$s are 0.52 and 0.34 for faithfulness and informativeness, respectively.}
    \label{tab:summ_human_eval}
    \vspace{-12pt}
\end{table}

\subsection{Faithfulness in Abstractive Summarization}\label{sec:result_summ}
The summarization results are shown in \autoref{tab:summ_results}. \TRIME{} improves faithfulness over the baseline on XSum, but the improvement is not clear on CNN/DM. In contrast, our \MODEL{} method greatly improves over the baseline, especially on DAE and P\textsubscript{ent}, which are specifically targeted towards hallucinations. Concretely, we improve FactCC by 0.91 points, DAE by 4.78 points, and P\textsubscript{ent} by 3 points on the XSum dataset. \MODEL{} improves the metrics on CNN/DM as well though to a smaller degree. This shows that allowing the model to pay specific attention to previous contexts in the input is helpful in reducing hallucinations.

We note that the ROUGE-L score for \MODEL{} is lower than the original model. This ROUGE-faithfulness tradeoff has been observed by many previous works \cite{chen-etal-2021-improving, kryscinski-etal-2020-evaluating,wan-bansal-2022-factpegasus,wan-etal-2023-faithfulness}, where the reference summary inherently contains hallucinations and thus does not overlap highly with the more faithful generated summaries.

To confirm this, we conduct a \textbf{human evaluation}. The results are shown in \autoref{tab:summ_human_eval}. \MODEL{} achieves the best faithfulness score, which is statistically significantly better than BART. This confirms our observation from automatic metric results in \autoref{tab:summ_results}. Though there is a small drop in informativeness, the difference between the three methods has no statistical significance.\footnote{We use boostrap test \cite{EfroTibs93} to determine statistical significance in our paper.} This shows that the drop in automated metrics such as ROUGE-L does not necessarily mean a decrease in informativeness. 

\subsection{Faithfulness in Data-to-Text Generation}\label{sec:result_data2text}
The results on LogicNLG are shown in \autoref{tab:logicnlg}. Similar to abstractive summarization, \MODEL{} can improve faithfulness on LogicNLG. Out of the four faithfulness metrics, \MODEL{} achieves the highest NLI-Acc, TAPEX-Acc, and TAPAS-Acc: \MODEL{} achieves  0.6 and 0.8 point improvements on TAPEX-Acc over BART and \TRIME{} respectively, and a 1.74 point improvement on TAPAS-Acc over the BART model. In the meantime, \MODEL{} obtains the best BLEU scores.

\section{Discussion and Conclusion}
In this work, we improve the context dependency of LMs by introducing a novel cache-LM training objective, \MODEL{}, which improves the existing cache-LM objective by adding an order-informed contrastive loss for the cache component.
On a synthetic dataset, we show that \MODEL{} is effective at retrieving the desired memories from the cache and breaking the softmax bottleneck. Furthermore, we demonstrate the effectiveness of \MODEL{} at improving the coherence of open-ended generation and improving faithfulness of abstractive summarization and data-to-text generation. 

We want to emphasize a couple of salient points with the recent trend of pushing for larger and more powerful models. Firstly, attention mechanisms alone cannot break the softmax bottleneck, as shown in \autoref{tab:ambiguous_llama}. Secondly, while increasing the model size can mitigate this bottleneck, the problem will persist unless we reach a size that truly encapsulates the complexity of human language. Cache-LM is a light alternative for breaking softmax bottleneck theoretically and improving context dependency empirically.

\begin{table}[!t]
    \centering
    \resizebox{\columnwidth}{!}{
    \begin{tabular}{l c c c c c }
    \toprule
    Model & BLEU-(1/2/3) & NA & SA & TX & TS \\
    \midrule
     BART & 56.27/37.07/25.63 & 85.46 & \textbf{53.45} & 63.97 & 63.74 \\
     \TRIME{} & 56.12/36.84/25.29 & 84.55 & 52.85 & 63.74 & 65.11  \\
     \MODEL{} & \textbf{56.65/37.56/26.25} & \textbf{85.67} & 53.12 & \textbf{64.58} & \textbf{65.48} \\
     \bottomrule
    \end{tabular}
    }
    \vspace{-5pt}
    \caption{Performance on LogicNLG (data-to-text generation) evaluated by BLEU scores, NLI-Acc (NA), SP-Acc (SA), TAPEX-Acc (TA), and TAPAS-Acc (TS). \MODEL{} improves over two baselines on BLEU and three faithfulness metrics: NA, TX, and TS.}
    \label{tab:logicnlg}
    \vspace{-12pt}
\end{table}

\section*{Acknowledgments}
We thank the reviewers and Haw-Shiuan Chang for helping with providing the Ambiguous Template data. This work was supported by NSF-CAREER Award 1846185, NSF-AI Engage Institute DRL-2112635, DARPA Machine Commonsense (MCS) Grant N66001-19-2-4031, and a Bloomberg Data Science Ph.D. Fellowship. The views contained in this article are those of the authors and not of the funding agency.

\section*{Limitations}
While we focus on the local memory to show that current LMs still benefit from better local context dependency, our method is also compatible with external memories, which can potentially further improve the performance of \MODEL{} in future work. We evaluate \MODEL{} using GPT2 and BART that at most consist of 774M parameters, which is smaller than the latest large LMs that can have billions of parameters. On the Ambiguous Template task, we do show that this problem exists for recent LLMs with LLaMA2 7B models and our method improves the cache alignment, but we hope that in the future we can explore scaling up the approach on large LMs to various tasks. We believe that our method is still helpful for larger models. But as larger models suffer less from softmax bottleneck (Section~\ref{sec:break_softmax}), how much it can help is an interesting problem to study in the future.
Another current limitation of this work is that due to the additional hyper-parameters (the $\lambda$ of the margin and the weight $\alpha$ of the contrastive loss), it becomes less straightforward to incorporate our \MODEL{} objective into pre-training compared to \TRIME{}. The training objective also considers that each token has a fixed margin (and thus assumes that each token is equally different), which can be improved by dynamically adjusting the margins. Although fine-tuning is cheaper and we show effective gains using \MODEL{} in fine-tuning, how to use \MODEL{} to pre-train LMs is also an interesting future work direction.

\section*{Ethical Considerations}
As the OpenAI team pointed out, GPT-2 does not distinguish fact from fiction, so it can not support use cases that require the generated text to be true.
In addition, GPT-2 reflects the biases inherent to the data they were trained on, so it can not be deployed unless the deployers first carry out a study of biases relevant to the intended use case. Though our \MODEL{} improves the coherence of GPT-2 generations, the above statement still holds. 
Similarly, despite that \MODEL{} improved the faithfulness of BART-large generations for abstractive summarization and data-to-text generation, such systems cannot be directly deployed and used in factuality-sensitive scenarios without further checks in place.

\bibliography{anthology,custom}

\begin{thebibliography}{60}
\expandafter\ifx\csname natexlab\endcsname\relax\def\natexlab#1{#1}\fi

\bibitem[{Alabdulkarim et~al.(2021)Alabdulkarim, Li, and
  Peng}]{alabdulkarim-etal-2021-automatic}
Amal Alabdulkarim, Siyan Li, and Xiangyu Peng. 2021.
\newblock \href {https://doi.org/10.18653/v1/2021.nuse-1.8} {Automatic story
  generation: Challenges and attempts}.
\newblock In \emph{Proceedings of the Third Workshop on Narrative
  Understanding}, pages 72--83, Virtual. Association for Computational
  Linguistics.

\bibitem[{Cao and Wang(2021)}]{cao-wang-2021-cliff}
Shuyang Cao and Lu~Wang. 2021.
\newblock \href {https://doi.org/10.18653/v1/2021.emnlp-main.532} {{CLIFF}:
  Contrastive learning for improving faithfulness and factuality in abstractive
  summarization}.
\newblock In \emph{Proceedings of the 2021 Conference on Empirical Methods in
  Natural Language Processing}, pages 6633--6649, Online and Punta Cana,
  Dominican Republic. Association for Computational Linguistics.

\bibitem[{Cao et~al.(2018)Cao, Wei, Li, and Li}]{cao2018faithful}
Ziqiang Cao, Furu Wei, Wenjie Li, and Sujian Li. 2018.
\newblock Faithful to the original: Fact-aware neural abstractive
  summarization.
\newblock In \emph{Proceedings of the Thirty-Second AAAI Conference on
  Artificial Intelligence and Thirtieth Innovative Applications of Artificial
  Intelligence Conference and Eighth AAAI Symposium on Educational Advances in
  Artificial Intelligence}, AAAI'18/IAAI'18/EAAI'18. AAAI Press.

\bibitem[{Chang and McCallum(2022)}]{chang-mccallum-2022-softmax}
Haw-Shiuan Chang and Andrew McCallum. 2022.
\newblock \href {https://doi.org/10.18653/v1/2022.acl-long.554} {Softmax
  bottleneck makes language models unable to represent multi-mode word
  distributions}.
\newblock In \emph{Proceedings of the 60th Annual Meeting of the Association
  for Computational Linguistics (Volume 1: Long Papers)}, pages 8048--8073,
  Dublin, Ireland. Association for Computational Linguistics.

\bibitem[{Chang et~al.(2023)Chang, Yao, Gon, Yu, and
  McCallum}]{chang2023revisiting}
Haw-Shiuan Chang, Zonghai Yao, Alolika Gon, Hong Yu, and Andrew McCallum. 2023.
\newblock Revisiting the architectures like pointer networks to efficiently
  improve the next word distribution, summarization factuality, and beyond.
\newblock In \emph{Findings of the Association for Computational Linguistics:
  ACL 2023 (Findings of ACL)}.

\bibitem[{Chen et~al.(2021)Chen, Zhang, Sone, and
  Roth}]{chen-etal-2021-improving}
Sihao Chen, Fan Zhang, Kazoo Sone, and Dan Roth. 2021.
\newblock \href {https://doi.org/10.18653/v1/2021.naacl-main.475} {Improving
  faithfulness in abstractive summarization with contrast candidate generation
  and selection}.
\newblock In \emph{Proceedings of the 2021 Conference of the North American
  Chapter of the Association for Computational Linguistics: Human Language
  Technologies}, pages 5935--5941, Online. Association for Computational
  Linguistics.

\bibitem[{Chen et~al.(2020{\natexlab{a}})Chen, Chen, Su, Chen, and
  Wang}]{chen-etal-2020-logical}
Wenhu Chen, Jianshu Chen, Yu~Su, Zhiyu Chen, and William~Yang Wang.
  2020{\natexlab{a}}.
\newblock \href {https://doi.org/10.18653/v1/2020.acl-main.708} {Logical
  natural language generation from open-domain tables}.
\newblock In \emph{Proceedings of the 58th Annual Meeting of the Association
  for Computational Linguistics}, pages 7929--7942, Online. Association for
  Computational Linguistics.

\bibitem[{Chen et~al.(2020{\natexlab{b}})Chen, Wang, Chen, Zhang, Wang, Li,
  Zhou, and Wang}]{Chen2020TabFact}
Wenhu Chen, Hongmin Wang, Jianshu Chen, Yunkai Zhang, Hong Wang, Shiyang Li,
  Xiyou Zhou, and William~Yang Wang. 2020{\natexlab{b}}.
\newblock \href {https://openreview.net/forum?id=rkeJRhNYDH} {Tabfact: A
  large-scale dataset for table-based fact verification}.
\newblock In \emph{International Conference on Learning Representations}.

\bibitem[{Efron and Tibshirani(1993)}]{EfroTibs93}
Bradley Efron and Robert~J. Tibshirani. 1993.
\newblock \emph{An Introduction to the Bootstrap}.
\newblock Number~57 in Monographs on Statistics and Applied Probability.
  Chapman \& Hall/CRC, Boca Raton, Florida, USA.

\bibitem[{Eisenschlos et~al.(2020)Eisenschlos, Krichene, and
  M{\"u}ller}]{eisenschlos-etal-2020-understanding}
Julian Eisenschlos, Syrine Krichene, and Thomas M{\"u}ller. 2020.
\newblock \href {https://doi.org/10.18653/v1/2020.findings-emnlp.27}
  {Understanding tables with intermediate pre-training}.
\newblock In \emph{Findings of the Association for Computational Linguistics:
  EMNLP 2020}, pages 281--296, Online. Association for Computational
  Linguistics.

\bibitem[{Fan et~al.(2018)Fan, Lewis, and Dauphin}]{fan-etal-2018-hierarchical}
Angela Fan, Mike Lewis, and Yann Dauphin. 2018.
\newblock \href {https://doi.org/10.18653/v1/P18-1082} {Hierarchical neural
  story generation}.
\newblock In \emph{Proceedings of the 56th Annual Meeting of the Association
  for Computational Linguistics (Volume 1: Long Papers)}, pages 889--898,
  Melbourne, Australia. Association for Computational Linguistics.

\bibitem[{Gao et~al.(2021)Gao, Yao, and Chen}]{gao-etal-2021-simcse}
Tianyu Gao, Xingcheng Yao, and Danqi Chen. 2021.
\newblock \href {https://doi.org/10.18653/v1/2021.emnlp-main.552} {{S}im{CSE}:
  Simple contrastive learning of sentence embeddings}.
\newblock In \emph{Proceedings of the 2021 Conference on Empirical Methods in
  Natural Language Processing}, pages 6894--6910, Online and Punta Cana,
  Dominican Republic. Association for Computational Linguistics.

\bibitem[{Goyal and Durrett(2021)}]{goyal-durrett-2021-annotating}
Tanya Goyal and Greg Durrett. 2021.
\newblock \href {https://doi.org/10.18653/v1/2021.naacl-main.114} {Annotating
  and modeling fine-grained factuality in summarization}.
\newblock In \emph{Proceedings of the 2021 Conference of the North American
  Chapter of the Association for Computational Linguistics: Human Language
  Technologies}, pages 1449--1462, Online. Association for Computational
  Linguistics.

\bibitem[{Grave et~al.(2017{\natexlab{a}})Grave, Cisse, and
  Joulin}]{grave2017unbounded}
Edouard Grave, Moustapha~M Cisse, and Armand Joulin. 2017{\natexlab{a}}.
\newblock Unbounded cache model for online language modeling with open
  vocabulary.
\newblock \emph{Advances in neural information processing systems}, 30.

\bibitem[{Grave et~al.(2017{\natexlab{b}})Grave, Joulin, and
  Usunier}]{grave-2017-improving}
Edouard Grave, Armand Joulin, and Nicolas Usunier. 2017{\natexlab{b}}.
\newblock \href {https://openreview.net/forum?id=B184E5qee} {Improving neural
  language models with a continuous cache}.
\newblock In \emph{International Conference on Learning Representations}.

\bibitem[{Grave et~al.(2017{\natexlab{c}})Grave, Joulin, and
  Usunier}]{grave2017improving}
Edouard Grave, Armand Joulin, and Nicolas Usunier. 2017{\natexlab{c}}.
\newblock \href {https://openreview.net/forum?id=B184E5qee} {Improving neural
  language models with a continuous cache}.
\newblock In \emph{International Conference on Learning Representations}.

\bibitem[{Gulcehre et~al.(2016)Gulcehre, Ahn, Nallapati, Zhou, and
  Bengio}]{gulcehre-etal-2016-pointing}
Caglar Gulcehre, Sungjin Ahn, Ramesh Nallapati, Bowen Zhou, and Yoshua Bengio.
  2016.
\newblock \href {https://doi.org/10.18653/v1/P16-1014} {Pointing the unknown
  words}.
\newblock In \emph{Proceedings of the 54th Annual Meeting of the Association
  for Computational Linguistics (Volume 1: Long Papers)}, pages 140--149,
  Berlin, Germany. Association for Computational Linguistics.

\bibitem[{Hermann et~al.(2015)Hermann, Kocisky, Grefenstette, Espeholt, Kay,
  Suleyman, and Blunsom}]{NIPS2015_afdec700}
Karl~Moritz Hermann, Tomas Kocisky, Edward Grefenstette, Lasse Espeholt, Will
  Kay, Mustafa Suleyman, and Phil Blunsom. 2015.
\newblock \href
  {https://proceedings.neurips.cc/paper_files/paper/2015/file/afdec7005cc9f14302cd0474fd0f3c96-Paper.pdf}
  {Teaching machines to read and comprehend}.
\newblock In \emph{Advances in Neural Information Processing Systems},
  volume~28. Curran Associates, Inc.

\bibitem[{Holtzman et~al.(2020)Holtzman, Buys, Du, Forbes, and
  Choi}]{Holtzman2020The}
Ari Holtzman, Jan Buys, Li~Du, Maxwell Forbes, and Yejin Choi. 2020.
\newblock \href {https://openreview.net/forum?id=rygGQyrFvH} {The curious case
  of neural text degeneration}.
\newblock In \emph{International Conference on Learning Representations}.

\bibitem[{Kanai et~al.(2018)Kanai, Fujiwara, Yamanaka, and Adachi}]{Sigsoftmax}
Sekitoshi Kanai, Yasuhiro Fujiwara, Yuki Yamanaka, and Shuichi Adachi. 2018.
\newblock \href
  {https://proceedings.neurips.cc/paper_files/paper/2018/file/9dcb88e0137649590b755372b040afad-Paper.pdf}
  {Sigsoftmax: Reanalysis of the softmax bottleneck}.
\newblock In \emph{Advances in Neural Information Processing Systems},
  volume~31. Curran Associates, Inc.

\bibitem[{Khandelwal et~al.(2020)Khandelwal, Levy, Jurafsky, Zettlemoyer, and
  Lewis}]{Khandelwal2020Generalization}
Urvashi Khandelwal, Omer Levy, Dan Jurafsky, Luke Zettlemoyer, and Mike Lewis.
  2020.
\newblock \href {https://openreview.net/forum?id=HklBjCEKvH} {Generalization
  through memorization: Nearest neighbor language models}.
\newblock In \emph{International Conference on Learning Representations}.

\bibitem[{Kryscinski et~al.(2020)Kryscinski, McCann, Xiong, and
  Socher}]{kryscinski-etal-2020-evaluating}
Wojciech Kryscinski, Bryan McCann, Caiming Xiong, and Richard Socher. 2020.
\newblock \href {https://doi.org/10.18653/v1/2020.emnlp-main.750} {Evaluating
  the factual consistency of abstractive text summarization}.
\newblock In \emph{Proceedings of the 2020 Conference on Empirical Methods in
  Natural Language Processing (EMNLP)}, pages 9332--9346, Online. Association
  for Computational Linguistics.

\bibitem[{Kuhn and De~Mori(1990)}]{kuhnvv1990cache}
Roland Kuhn and Renato De~Mori. 1990.
\newblock A cache-based natural language model for speech recognition.
\newblock \emph{IEEE transactions on pattern analysis and machine
  intelligence}, 12(6):570--583.

\bibitem[{Lewis et~al.(2020)Lewis, Liu, Goyal, Ghazvininejad, Mohamed, Levy,
  Stoyanov, and Zettlemoyer}]{lewis-etal-2020-bart}
Mike Lewis, Yinhan Liu, Naman Goyal, Marjan Ghazvininejad, Abdelrahman Mohamed,
  Omer Levy, Veselin Stoyanov, and Luke Zettlemoyer. 2020.
\newblock \href {https://doi.org/10.18653/v1/2020.acl-main.703} {{BART}:
  Denoising sequence-to-sequence pre-training for natural language generation,
  translation, and comprehension}.
\newblock In \emph{Proceedings of the 58th Annual Meeting of the Association
  for Computational Linguistics}, pages 7871--7880, Online. Association for
  Computational Linguistics.

\bibitem[{Lhoest et~al.(2021)Lhoest, Villanova~del Moral, Jernite, Thakur, von
  Platen, Patil, Chaumond, Drame, Plu, Tunstall, Davison, {\v{S}}a{\v{s}}ko,
  Chhablani, Malik, Brandeis, Le~Scao, Sanh, Xu, Patry, McMillan-Major, Schmid,
  Gugger, Delangue, Matussi{\`e}re, Debut, Bekman, Cistac, Goehringer, Mustar,
  Lagunas, Rush, and Wolf}]{lhoest-etal-2021-datasets}
Quentin Lhoest, Albert Villanova~del Moral, Yacine Jernite, Abhishek Thakur,
  Patrick von Platen, Suraj Patil, Julien Chaumond, Mariama Drame, Julien Plu,
  Lewis Tunstall, Joe Davison, Mario {\v{S}}a{\v{s}}ko, Gunjan Chhablani,
  Bhavitvya Malik, Simon Brandeis, Teven Le~Scao, Victor Sanh, Canwen Xu,
  Nicolas Patry, Angelina McMillan-Major, Philipp Schmid, Sylvain Gugger,
  Cl{\'e}ment Delangue, Th{\'e}o Matussi{\`e}re, Lysandre Debut, Stas Bekman,
  Pierric Cistac, Thibault Goehringer, Victor Mustar, Fran{\c{c}}ois Lagunas,
  Alexander Rush, and Thomas Wolf. 2021.
\newblock \href {https://doi.org/10.18653/v1/2021.emnlp-demo.21} {Datasets: A
  community library for natural language processing}.
\newblock In \emph{Proceedings of the 2021 Conference on Empirical Methods in
  Natural Language Processing: System Demonstrations}, pages 175--184, Online
  and Punta Cana, Dominican Republic. Association for Computational
  Linguistics.

\bibitem[{Li et~al.(2022{\natexlab{a}})Li, Holtzman, Fried, Liang, Eisner,
  Hashimoto, Zettlemoyer, and Lewis}]{contrastive_decoding}
Xiang~Lisa Li, Ari Holtzman, Daniel Fried, Percy Liang, Jason Eisner, Tatsunori
  Hashimoto, Luke Zettlemoyer, and Mike Lewis. 2022{\natexlab{a}}.
\newblock \href {https://doi.org/10.48550/ARXIV.2210.15097} {Contrastive
  decoding: Open-ended text generation as optimization}.

\bibitem[{Li et~al.(2022{\natexlab{b}})Li, Holtzman, Fried, Liang, Eisner,
  Hashimoto, Zettlemoyer, and Lewis}]{li2022contrastive}
Xiang~Lisa Li, Ari Holtzman, Daniel Fried, Percy Liang, Jason Eisner, Tatsunori
  Hashimoto, Luke Zettlemoyer, and Mike Lewis. 2022{\natexlab{b}}.
\newblock \href {http://arxiv.org/abs/2210.15097} {Contrastive decoding:
  Open-ended text generation as optimization}.

\bibitem[{Lin(2004)}]{lin-2004-rouge}
Chin-Yew Lin. 2004.
\newblock \href {https://aclanthology.org/W04-1013} {{ROUGE}: A package for
  automatic evaluation of summaries}.
\newblock In \emph{Text Summarization Branches Out}, pages 74--81, Barcelona,
  Spain. Association for Computational Linguistics.

\bibitem[{Liu et~al.(2022{\natexlab{a}})Liu, Dong, Okazaki, Han, and
  Zhang}]{liu-etal-2022-plog}
Ao~Liu, Haoyu Dong, Naoaki Okazaki, Shi Han, and Dongmei Zhang.
  2022{\natexlab{a}}.
\newblock \href {https://aclanthology.org/2022.emnlp-main.373} {{PLOG}:
  Table-to-logic pretraining for logical table-to-text generation}.
\newblock In \emph{Proceedings of the 2022 Conference on Empirical Methods in
  Natural Language Processing}, pages 5531--5546, Abu Dhabi, United Arab
  Emirates. Association for Computational Linguistics.

\bibitem[{Liu et~al.(2022{\natexlab{b}})Liu, Chen, Guo, Ziyadi, Lin, Chen, and
  Lou}]{liu2022tapex}
Qian Liu, Bei Chen, Jiaqi Guo, Morteza Ziyadi, Zeqi Lin, Weizhu Chen, and
  Jian-Guang Lou. 2022{\natexlab{b}}.
\newblock \href {https://openreview.net/forum?id=O50443AsCP} {{TAPEX}: Table
  pre-training via learning a neural {SQL} executor}.
\newblock In \emph{International Conference on Learning Representations}.

\bibitem[{Liu et~al.(2022{\natexlab{c}})Liu, Liu, Radev, and
  Neubig}]{liu-etal-2022-brio}
Yixin Liu, Pengfei Liu, Dragomir Radev, and Graham Neubig. 2022{\natexlab{c}}.
\newblock \href {https://doi.org/10.18653/v1/2022.acl-long.207} {{BRIO}:
  Bringing order to abstractive summarization}.
\newblock In \emph{Proceedings of the 60th Annual Meeting of the Association
  for Computational Linguistics (Volume 1: Long Papers)}, pages 2890--2903,
  Dublin, Ireland. Association for Computational Linguistics.

\bibitem[{Loshchilov and Hutter(2019)}]{loshchilov2018decoupled}
Ilya Loshchilov and Frank Hutter. 2019.
\newblock \href {https://openreview.net/forum?id=Bkg6RiCqY7} {Decoupled weight
  decay regularization}.
\newblock In \emph{International Conference on Learning Representations}.

\bibitem[{Maynez et~al.(2020)Maynez, Narayan, Bohnet, and
  McDonald}]{maynez-etal-2020-faithfulness}
Joshua Maynez, Shashi Narayan, Bernd Bohnet, and Ryan McDonald. 2020.
\newblock \href {https://doi.org/10.18653/v1/2020.acl-main.173} {On
  faithfulness and factuality in abstractive summarization}.
\newblock In \emph{Proceedings of the 58th Annual Meeting of the Association
  for Computational Linguistics}, pages 1906--1919, Online. Association for
  Computational Linguistics.

\bibitem[{Meister et~al.(2022)Meister, Pimentel, Wiher, and
  Cotterell}]{meister+al.pre22}
Clara Meister, Tiago Pimentel, Gian Wiher, and Ryan Cotterell. 2022.
\newblock \href {http://arxiv.org/abs/2202.00666} {Locally typical sampling}.
\newblock \emph{Transactions of the Association for Computational Linguistics},
  abs/2202.00666.

\bibitem[{Merity et~al.(2017)Merity, Xiong, Bradbury, and
  Socher}]{merity2017pointer}
Stephen Merity, Caiming Xiong, James Bradbury, and Richard Socher. 2017.
\newblock \href {https://openreview.net/forum?id=Byj72udxe} {Pointer sentinel
  mixture models}.
\newblock In \emph{International Conference on Learning Representations}.

\bibitem[{Mikolov et~al.(2013)Mikolov, Sutskever, Chen, Corrado, and
  Dean}]{NIPS2013_9aa42b31}
Tomas Mikolov, Ilya Sutskever, Kai Chen, Greg~S Corrado, and Jeff Dean. 2013.
\newblock \href
  {https://proceedings.neurips.cc/paper_files/paper/2013/file/9aa42b31882ec039965f3c4923ce901b-Paper.pdf}
  {Distributed representations of words and phrases and their
  compositionality}.
\newblock In \emph{Advances in Neural Information Processing Systems},
  volume~26. Curran Associates, Inc.

\bibitem[{Min et~al.(2022)Min, Shi, Lewis, Chen, tau Yih, Hajishirzi, and
  Zettlemoyer}]{min2022nonparametric}
Sewon Min, Weijia Shi, Mike Lewis, Xilun Chen, Wen tau Yih, Hannaneh
  Hajishirzi, and Luke Zettlemoyer. 2022.
\newblock \href {http://arxiv.org/abs/2212.01349} {Nonparametric masked
  language modeling}.

\bibitem[{Nan et~al.(2021)Nan, Nallapati, Wang, Nogueira~dos Santos, Zhu,
  Zhang, McKeown, and Xiang}]{nan-etal-2021-entity}
Feng Nan, Ramesh Nallapati, Zhiguo Wang, Cicero Nogueira~dos Santos, Henghui
  Zhu, Dejiao Zhang, Kathleen McKeown, and Bing Xiang. 2021.
\newblock \href {https://doi.org/10.18653/v1/2021.eacl-main.235} {Entity-level
  factual consistency of abstractive text summarization}.
\newblock In \emph{Proceedings of the 16th Conference of the European Chapter
  of the Association for Computational Linguistics: Main Volume}, pages
  2727--2733, Online. Association for Computational Linguistics.

\bibitem[{Narayan et~al.(2018)Narayan, Cohen, and
  Lapata}]{narayan-etal-2018-dont}
Shashi Narayan, Shay~B. Cohen, and Mirella Lapata. 2018.
\newblock \href {https://doi.org/10.18653/v1/D18-1206} {Don{'}t give me the
  details, just the summary! topic-aware convolutional neural networks for
  extreme summarization}.
\newblock In \emph{Proceedings of the 2018 Conference on Empirical Methods in
  Natural Language Processing}, pages 1797--1807, Brussels, Belgium.
  Association for Computational Linguistics.

\bibitem[{Papineni et~al.(2002)Papineni, Roukos, Ward, and
  Zhu}]{papineni-etal-2002-bleu}
Kishore Papineni, Salim Roukos, Todd Ward, and Wei-Jing Zhu. 2002.
\newblock \href {https://doi.org/10.3115/1073083.1073135} {{B}leu: a method for
  automatic evaluation of machine translation}.
\newblock In \emph{Proceedings of the 40th Annual Meeting of the Association
  for Computational Linguistics}, pages 311--318, Philadelphia, Pennsylvania,
  USA. Association for Computational Linguistics.

\bibitem[{Pillutla et~al.(2021)Pillutla, Swayamdipta, Zellers, Thickstun,
  Welleck, Choi, and Harchaoui}]{NEURIPS2021_260c2432}
Krishna Pillutla, Swabha Swayamdipta, Rowan Zellers, John Thickstun, Sean
  Welleck, Yejin Choi, and Zaid Harchaoui. 2021.
\newblock \href
  {https://proceedings.neurips.cc/paper_files/paper/2021/file/260c2432a0eecc28ce03c10dadc078a4-Paper.pdf}
  {Mauve: Measuring the gap between neural text and human text using divergence
  frontiers}.
\newblock In \emph{Advances in Neural Information Processing Systems},
  volume~34, pages 4816--4828. Curran Associates, Inc.

\bibitem[{Radford et~al.(2019)Radford, Wu, Child, Luan, Amodei, and
  Sutskever}]{radford2019language}
Alec Radford, Jeff Wu, Rewon Child, David Luan, Dario Amodei, and Ilya
  Sutskever. 2019.
\newblock Language models are unsupervised multitask learners.
\newblock \emph{OpenAI blog}.

\bibitem[{Rashkin et~al.(2020)Rashkin, Celikyilmaz, Choi, and
  Gao}]{rashkin-etal-2020-plotmachines}
Hannah Rashkin, Asli Celikyilmaz, Yejin Choi, and Jianfeng Gao. 2020.
\newblock \href {https://doi.org/10.18653/v1/2020.emnlp-main.349}
  {{P}lot{M}achines: Outline-conditioned generation with dynamic plot state
  tracking}.
\newblock In \emph{Proceedings of the 2020 Conference on Empirical Methods in
  Natural Language Processing (EMNLP)}, pages 4274--4295, Online. Association
  for Computational Linguistics.

\bibitem[{Reimers and Gurevych(2019)}]{reimers-gurevych-2019-sentence}
Nils Reimers and Iryna Gurevych. 2019.
\newblock \href {https://doi.org/10.18653/v1/D19-1410} {Sentence-{BERT}:
  Sentence embeddings using {S}iamese {BERT}-networks}.
\newblock In \emph{Proceedings of the 2019 Conference on Empirical Methods in
  Natural Language Processing and the 9th International Joint Conference on
  Natural Language Processing (EMNLP-IJCNLP)}, pages 3982--3992, Hong Kong,
  China. Association for Computational Linguistics.

\bibitem[{See et~al.(2017)See, Liu, and Manning}]{see2017get}
Abigail See, Peter~J Liu, and Christopher~D Manning. 2017.
\newblock Get to the point: Summarization with pointer-generator networks.
\newblock In \emph{Proceedings of the 55th Annual Meeting of the Association
  for Computational Linguistics (Volume 1: Long Papers)}, pages 1073--1083.

\bibitem[{Su et~al.(2022)Su, Lan, Wang, Yogatama, Kong, and Collier}]{su2022a}
Yixuan Su, Tian Lan, Yan Wang, Dani Yogatama, Lingpeng Kong, and Nigel Collier.
  2022.
\newblock \href {https://openreview.net/forum?id=V88BafmH9Pj} {A contrastive
  framework for neural text generation}.
\newblock In \emph{Advances in Neural Information Processing Systems}.

\bibitem[{Thomson and Reiter(2020)}]{thomson-reiter-2020-gold}
Craig Thomson and Ehud Reiter. 2020.
\newblock \href {https://aclanthology.org/2020.inlg-1.22} {A gold standard
  methodology for evaluating accuracy in data-to-text systems}.
\newblock In \emph{Proceedings of the 13th International Conference on Natural
  Language Generation}, pages 158--168, Dublin, Ireland. Association for
  Computational Linguistics.

\bibitem[{Touvron et~al.(2023)Touvron, Martin, Stone, Albert, Almahairi,
  Babaei, Bashlykov, Batra, Bhargava, Bhosale, Bikel, Blecher, Ferrer, Chen,
  Cucurull, Esiobu, Fernandes, Fu, Fu, Fuller, Gao, Goswami, Goyal, Hartshorn,
  Hosseini, Hou, Inan, Kardas, Kerkez, Khabsa, Kloumann, Korenev, Koura,
  Lachaux, Lavril, Lee, Liskovich, Lu, Mao, Martinet, Mihaylov, Mishra,
  Molybog, Nie, Poulton, Reizenstein, Rungta, Saladi, Schelten, Silva, Smith,
  Subramanian, Tan, Tang, Taylor, Williams, Kuan, Xu, Yan, Zarov, Zhang, Fan,
  Kambadur, Narang, Rodriguez, Stojnic, Edunov, and Scialom}]{touvron2023llama}
Hugo Touvron, Louis Martin, Kevin Stone, Peter Albert, Amjad Almahairi, Yasmine
  Babaei, Nikolay Bashlykov, Soumya Batra, Prajjwal Bhargava, Shruti Bhosale,
  Dan Bikel, Lukas Blecher, Cristian~Canton Ferrer, Moya Chen, Guillem
  Cucurull, David Esiobu, Jude Fernandes, Jeremy Fu, Wenyin Fu, Brian Fuller,
  Cynthia Gao, Vedanuj Goswami, Naman Goyal, Anthony Hartshorn, Saghar
  Hosseini, Rui Hou, Hakan Inan, Marcin Kardas, Viktor Kerkez, Madian Khabsa,
  Isabel Kloumann, Artem Korenev, Punit~Singh Koura, Marie-Anne Lachaux,
  Thibaut Lavril, Jenya Lee, Diana Liskovich, Yinghai Lu, Yuning Mao, Xavier
  Martinet, Todor Mihaylov, Pushkar Mishra, Igor Molybog, Yixin Nie, Andrew
  Poulton, Jeremy Reizenstein, Rashi Rungta, Kalyan Saladi, Alan Schelten, Ruan
  Silva, Eric~Michael Smith, Ranjan Subramanian, Xiaoqing~Ellen Tan, Binh Tang,
  Ross Taylor, Adina Williams, Jian~Xiang Kuan, Puxin Xu, Zheng Yan, Iliyan
  Zarov, Yuchen Zhang, Angela Fan, Melanie Kambadur, Sharan Narang, Aurelien
  Rodriguez, Robert Stojnic, Sergey Edunov, and Thomas Scialom. 2023.
\newblock \href {http://arxiv.org/abs/2307.09288} {Llama 2: Open foundation and
  fine-tuned chat models}.

\bibitem[{van~den Oord et~al.(2019)van~den Oord, Li, and
  Vinyals}]{oord2019representation}
Aaron van~den Oord, Yazhe Li, and Oriol Vinyals. 2019.
\newblock \href {http://arxiv.org/abs/1807.03748} {Representation learning with
  contrastive predictive coding}.

\bibitem[{Vaswani et~al.(2017)Vaswani, Shazeer, Parmar, Uszkoreit, Jones,
  Gomez, Kaiser, and Polosukhin}]{transformer}
Ashish Vaswani, Noam Shazeer, Niki Parmar, Jakob Uszkoreit, Llion Jones,
  Aidan~N Gomez, \L~ukasz Kaiser, and Illia Polosukhin. 2017.
\newblock \href
  {https://proceedings.neurips.cc/paper_files/paper/2017/file/3f5ee243547dee91fbd053c1c4a845aa-Paper.pdf}
  {Attention is all you need}.
\newblock In \emph{Advances in Neural Information Processing Systems},
  volume~30. Curran Associates, Inc.

\bibitem[{Vinyals et~al.(2015)Vinyals, Fortunato, and
  Jaitly}]{vinyals2015pointer}
Oriol Vinyals, Meire Fortunato, and Navdeep Jaitly. 2015.
\newblock Pointer networks.
\newblock \emph{Advances in neural information processing systems}, 28.

\bibitem[{Wan and Bansal(2022)}]{wan-bansal-2022-factpegasus}
David Wan and Mohit Bansal. 2022.
\newblock \href {https://doi.org/10.18653/v1/2022.naacl-main.74}
  {{F}act{PEGASUS}: Factuality-aware pre-training and fine-tuning for
  abstractive summarization}.
\newblock In \emph{Proceedings of the 2022 Conference of the North American
  Chapter of the Association for Computational Linguistics: Human Language
  Technologies}, pages 1010--1028, Seattle, United States. Association for
  Computational Linguistics.

\bibitem[{Wan et~al.(2023)Wan, Liu, McKeown, Dreyer, and
  Bansal}]{wan-etal-2023-faithfulness}
David Wan, Mengwen Liu, Kathleen McKeown, Markus Dreyer, and Mohit Bansal.
  2023.
\newblock \href {https://aclanthology.org/2023.eacl-main.210}
  {Faithfulness-aware decoding strategies for abstractive summarization}.
\newblock In \emph{Proceedings of the 17th Conference of the European Chapter
  of the Association for Computational Linguistics}, pages 2864--2880,
  Dubrovnik, Croatia. Association for Computational Linguistics.

\bibitem[{Weng et~al.(2020)Weng, Yu, Wei, and Luo}]{weng-etal-2020-towards}
Rongxiang Weng, Heng Yu, Xiangpeng Wei, and Weihua Luo. 2020.
\newblock \href {https://doi.org/10.18653/v1/2020.emnlp-main.212} {Towards
  enhancing faithfulness for neural machine translation}.
\newblock In \emph{Proceedings of the 2020 Conference on Empirical Methods in
  Natural Language Processing (EMNLP)}, pages 2675--2684, Online. Association
  for Computational Linguistics.

\bibitem[{Wolf et~al.(2020)Wolf, Debut, Sanh, Chaumond, Delangue, Moi, Cistac,
  Rault, Louf, Funtowicz, Davison, Shleifer, von Platen, Ma, Jernite, Plu, Xu,
  Le~Scao, Gugger, Drame, Lhoest, and Rush}]{wolf-etal-2020-transformers}
Thomas Wolf, Lysandre Debut, Victor Sanh, Julien Chaumond, Clement Delangue,
  Anthony Moi, Pierric Cistac, Tim Rault, Remi Louf, Morgan Funtowicz, Joe
  Davison, Sam Shleifer, Patrick von Platen, Clara Ma, Yacine Jernite, Julien
  Plu, Canwen Xu, Teven Le~Scao, Sylvain Gugger, Mariama Drame, Quentin Lhoest,
  and Alexander Rush. 2020.
\newblock \href {https://doi.org/10.18653/v1/2020.emnlp-demos.6} {Transformers:
  State-of-the-art natural language processing}.
\newblock In \emph{Proceedings of the 2020 Conference on Empirical Methods in
  Natural Language Processing: System Demonstrations}, pages 38--45, Online.
  Association for Computational Linguistics.

\bibitem[{Yang et~al.(2018)Yang, Dai, Salakhutdinov, and
  Cohen}]{yang2018breaking}
Zhilin Yang, Zihang Dai, Ruslan Salakhutdinov, and William~W. Cohen. 2018.
\newblock \href {https://openreview.net/forum?id=HkwZSG-CZ} {Breaking the
  softmax bottleneck: A high-rank {RNN} language model}.
\newblock In \emph{International Conference on Learning Representations}.

\bibitem[{Yang et~al.(2019)Yang, Luong, Salakhutdinov, and Le}]{Mixtape}
Zhilin Yang, Thang Luong, Russ~R Salakhutdinov, and Quoc~V Le. 2019.
\newblock \href
  {https://proceedings.neurips.cc/paper_files/paper/2019/file/512fc3c5227f637e41437c999a2d3169-Paper.pdf}
  {Mixtape: Breaking the softmax bottleneck efficiently}.
\newblock In \emph{Advances in Neural Information Processing Systems},
  volume~32. Curran Associates, Inc.

\bibitem[{Yin et~al.(2020)Yin, Neubig, Yih, and Riedel}]{yin-etal-2020-tabert}
Pengcheng Yin, Graham Neubig, Wen-tau Yih, and Sebastian Riedel. 2020.
\newblock \href {https://doi.org/10.18653/v1/2020.acl-main.745} {{T}a{BERT}:
  Pretraining for joint understanding of textual and tabular data}.
\newblock In \emph{Proceedings of the 58th Annual Meeting of the Association
  for Computational Linguistics}, pages 8413--8426, Online. Association for
  Computational Linguistics.

\bibitem[{Yogatama et~al.(2021)Yogatama, de~Masson~d{'}Autume, and
  Kong}]{yogatama-etal-2021-adaptive}
Dani Yogatama, Cyprien de~Masson~d{'}Autume, and Lingpeng Kong. 2021.
\newblock \href {https://doi.org/10.1162/tacl_a_00371} {Adaptive semiparametric
  language models}.
\newblock \emph{Transactions of the Association for Computational Linguistics},
  9:362--373.

\bibitem[{Zhong et~al.(2022)Zhong, Lei, and Chen}]{zhong-etal-2022-training}
Zexuan Zhong, Tao Lei, and Danqi Chen. 2022.
\newblock \href {https://aclanthology.org/2022.emnlp-main.382} {Training
  language models with memory augmentation}.
\newblock In \emph{Proceedings of the 2022 Conference on Empirical Methods in
  Natural Language Processing}, pages 5657--5673, Abu Dhabi, United Arab
  Emirates. Association for Computational Linguistics.

\end{thebibliography}
\bibliographystyle{acl_natbib}

\appendix

\begin{table*}
    \resizebox{\textwidth}{!}{
    \begin{tabular}{c c c c c c c c }
    \toprule
    Task & Dataset & Learning rate & Steps/Epochs & Warmup ratio & Batch size & Max input tokens & Max output tokens \\
    \midrule
    \multicolumn{2}{c}{Ambiguous Template} & 1e-5 & 5 ep. & 0.0 & 256 & - & 1024 \\
    Open-ended Generation & WritingPrompt & 5e-5 & 3 ep. & 0.2 & 512 & - & 1024 \\
    \multirow{2}{*}{Summarization}  & XSum & 5e-5 & 15,000 steps & 0.0 & 128 & 512 & 64 \\
    & CNN/DM & 3e-5 & 20,000 steps & 0.0 & 128 & 512 & 128 \\
    Data-to-text & LogicNLG & 5e-5 & 10 eps & 0.0 & 64 & 500 & 200\\ 
    \bottomrule
    \end{tabular}
    }
    \caption{Hyper-parameters for all tasks.}
    \label{tab:hyperparams}
\end{table*}

\begin{table*}[!t]
    \small
    \begin{tabularx}{\textwidth}{p{0.05\textwidth}p{0.1\textwidth}|p{0.36\textwidth}|p{0.36\textwidth}}
    \toprule
    & Context & \emph{Houston} and \emph{Pennsylvania} are my favorites, and I especially love & The \emph{brother} and the \emph{granddaughter} are my favorites, and I especially love the\\
    \midrule
    \multirow{3}{*}{Full} & GPT2-large & \emph{Pennsylvania}, Harris, Pittsburgh &   \emph{brother}, niece, \emph{granddaughter} \\
    & \TRIME{} & \emph{Pennsylvania}, Philadelphia, Pittsburgh & \emph{brother}, niece, \emph{granddaughter} \\
    & \MODEL{} & \emph{Pennsylvania}, \emph{Houston}, Harris & \emph{brother}, \emph{granddaughter}, niece \\
    \midrule
    \multirow{3}{*}{\parbox{1cm}{Cache-\\only}} & GPT2-large & and, I, \emph{Houston} & the, \emph{granddaughter}, I \\
    & \TRIME{}  & I, and, \emph{Houston} & the, \emph{granddaughter}, favorites \\
    & \MODEL{} &  \emph{Pennsylvania}, \emph{Houston}, and & \emph{granddaughter}, \emph{brother}, the \\
    \bottomrule
    \end{tabularx}
    \caption{Qualitative examples from Ambiguous Template. For both the full and cache-only settings, \MODEL{} retrieves the correct two tokens from the context as the top predictions.}
    \label{tab:amb_qualitative}
\end{table*}

\section{Experimental Setup Details}
\label{sec:experimental_appendix}
Unless specified, we use Huggingface's Transformers library~\cite{wolf-etal-2020-transformers} to train the models. We use the trainer's default setting, including AdamW optimizer~\cite{loshchilov2018decoupled} and a linear rate scheduler. We use mixed precision and deepspeed. We use RTX A6000 GPUs with 48GB memory and A100 GPUs with 80GB memory.

For hyperparameter tuning, we try learning rate of \{1e-5,3e-5,5e-5\} and $\lambda$ between \{0.001,0.0001,0.00001\}, and contrastive weight \{0.5,1.0\} for all tasks. For \MODEL{}, we use $\lambda$ of 0.001 and the contrastive weight $\alpha=1$, unless otherwise specified.

\subsection{Ambiguous Template}
The dataset consists of 122k, 250k, and 122k examples for train, dev, and test sets, respectively. The test set has no overlap of diagonal words with the training set. Following \citet{chang-mccallum-2022-softmax}, we freeze output vocab to prevent overfitting, and get loss only from the last token (the target token). We select the model based on validation loss. For GPT2-large-based models, training the original model took around 30 minutes, \TRIME{} and \MODEL{} took around an hour with 4 RTX A6000. On GPT2-small-based models, training took 10 minutes, 15 minutes, and 15 minutes, for orig, \TRIME{}, and \MODEL{}, respectively.

\subsection{Prompt Continuation}
WritingPrompts\footnote{\url{https://github.com/urvashik/knnmt/blob/master/examples/stories/README.md}} \cite{fan-etal-2018-hierarchical} contain 273k, 16k, and 15k examples in the train, dev, and test sets. We use the full train and dev sets, while we sample 5000 examples from the test set for final evaluation to save time. 
We first train the models using the different objectives on the training set. We split the text into blocks of 512 tokens. For generation, we decode with nucleus sampling with $p=0.95$ and three random seeds=\{0,1,42\}, and average the scores. Training the original small model takes around 1.5 hours, \TRIME{} takes around 2 hours, and \MODEL{} takes around 3 hours. Training GPT-2 large, \TRIME{} and \MODEL{} takes around 6 hours, 7 hours, and 11.5 hours on 2 A100s, respectively.

\subsection{Summarization}
XSum is a news summarization dataset consisting of BBC articles and contains 204k/11k/11k examples in the train/dev/test set. CNN/DM consists of Dailymail and CNN articles and the dataset consists of 287k/13k/11k examples in the train/dev/test set.
We use the official packages for the faithfulness metrics.\footnote{FactCC: \url{https://github.com/salesforce/factCC}. DAE: \url{https://github.com/tagoyal/factuality-datasets}.} We calculate P\textsubscript{ent} by using spacy to extract entities and only consider 
[PERSON, FAC, GPE, ORG, NORP, LOC, EVENT] as the allowed entity types. We use Huggingface's Dataset library~\cite{lhoest-etal-2021-datasets} for loading the XSum \cite{narayan-etal-2018-dont} and CNN/DM \cite{NIPS2015_afdec700} datasets. And we use Huggingface's Metrics library for calculating ROUGE scores. Training the original model, \TRIME{}, and \MODEL{} all took around 5 hours for XSum and training orig, \TRIME{} and \MODEL{} all took around 4 hours for CNN/DM on 4 A6000s. 

\subsection{Data-to-text}\label{sec:experimental_data2text_appendix}
We follow \citet{liu-etal-2022-plog} for pre-processing dataset, such as adding numerical pre-computation to the tables. We use a contrastive weight $\alpha=0.5$. LogicNLG \cite{chen-etal-2020-logical} consists of 28k training, 4k validation, and 4k test examples. We  use original evaluation scripts for the faithfulness metrics, and the BLEU calculation script provided by the original dataset.\footnote{\url{https://github.com/wenhuchen/LogicNLG/blob/master/evaluate.py}} Training the original model and \TRIME{} took 2 hours, and \MODEL{} took around 5 hours on 2 A100.

\section{Human Evaluation Details}
For both human evaluations, we use Amazon Mechanical Turk to do the annotation. We have the same set of requirements: The workers need to be from the United States, have more than 10,000 number of HITS approved, and an approval rate greater than 98\%.

\subsection{Open-ended Generation}\label{sec:prompt_human}
We use Amazon Mechanical Turk to annotate whether human prefers the continuation by \TRIME{} or by \MODEL{}. We  do not include the original model, since \TRIME{} shows better performance on the automatic metrics. We select examples where the difference between their characters is less than 200 characters to ensure that the length is similar (since shorter texts will naturally be more coherent). We collect 3 annotations per example for 100 randomly selected examples, yielding 300 annotations. We take the percentage of passages that are judged as coherent and/or fluent. 

\begin{figure*}
    \includegraphics[width=1.0\textwidth]{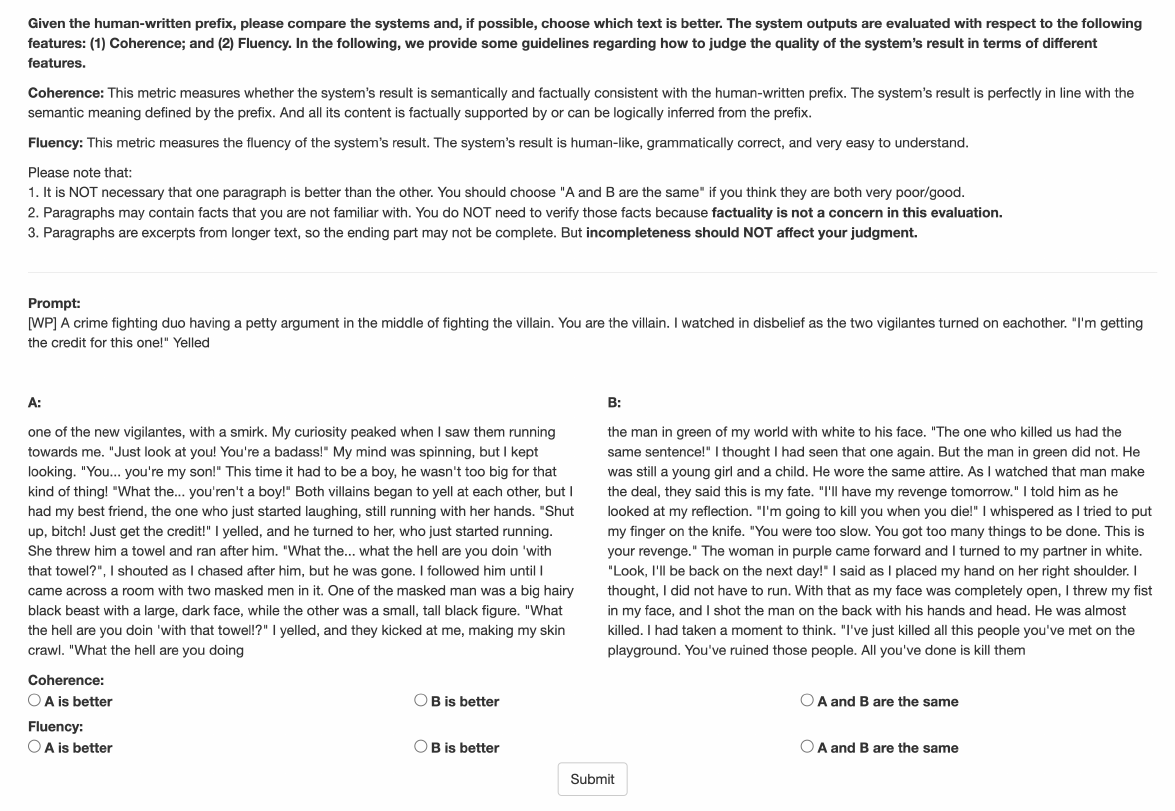}
    \caption{Human annotation page for evaluating coherence and fluency for prompt continuation.}
    \label{fig:human_eval_open}
\end{figure*}

We pay 0.5 USD per HIT, and the average time it takes is around 2.5 minutes, which yields an hourly rate of $\geq \$12$ per hour. An example of the annotation page is shown in \autoref{fig:human_eval_open}.

\subsection{Summarization}\label{sec:summ_human}
We follow the same setup as \citet{wan-bansal-2022-factpegasus}, and also use a qualification test where we rate the faithfulness of the selected generated summaries. Only workers with the correct annotation can perform the actual task.

We select the most important sentences and replace the less relevant sentences with an ellipsis to reduce the overload for the workers. We select ten most relevant sentences from the document by cosine similarity of the sentence embedding using SentenceTransformer\footnote{We use the all-mpnet-base-v2  model.} \cite{reimers-gurevych-2019-sentence} for each summary and then combine and show all the selected relevant sentences from each summary.
                    
\begin{figure*}
    \includegraphics[width=1.0\textwidth]{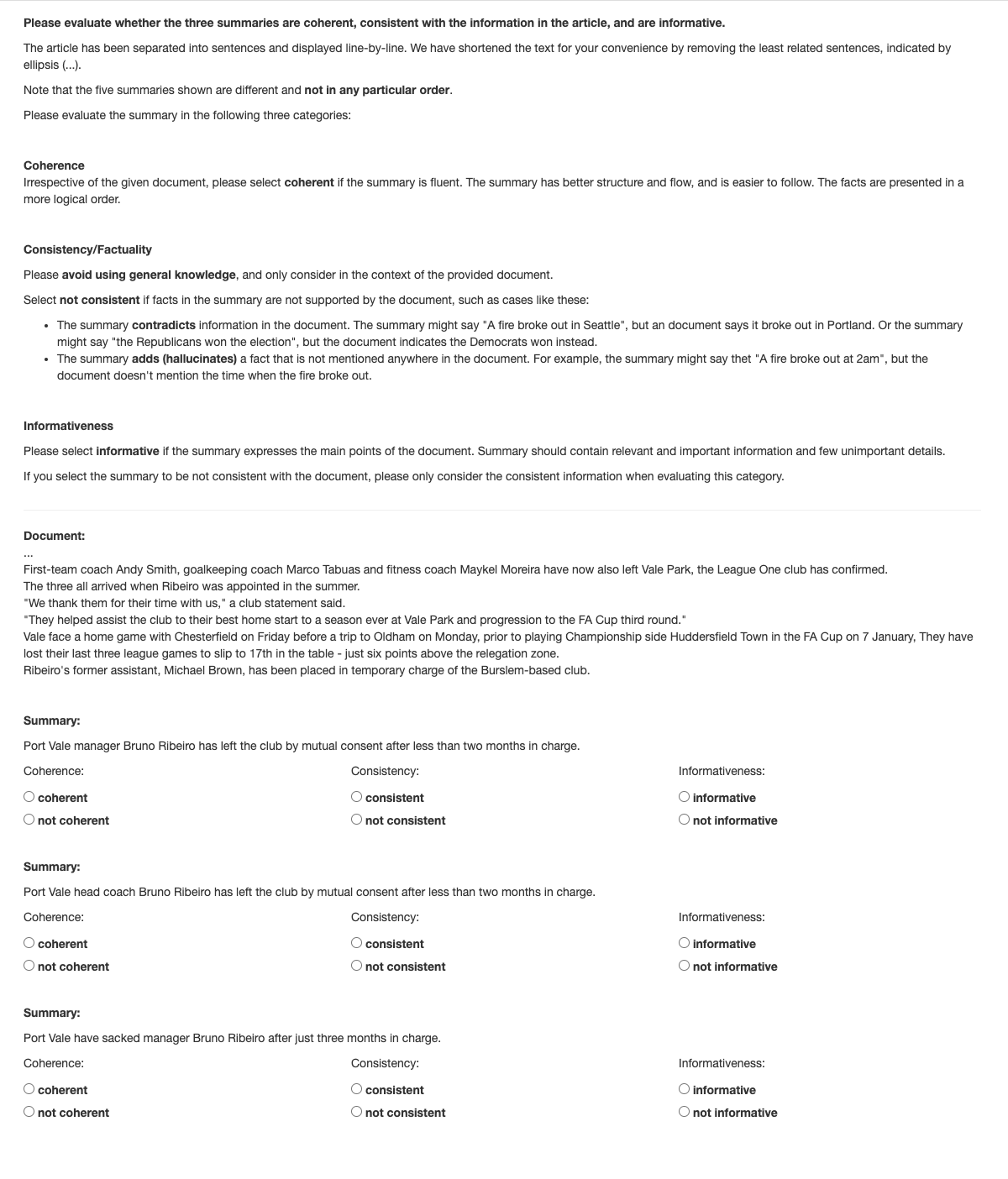}
    \caption{Human annotation page for evaluating faithfulness and informativeness for abstractive summarization.}
    \label{fig:human_eval_summ}
\end{figure*}

Each task consists of three unique workers, where we take the mean as the scores for this document. The final score is the mean factuality score across all documents. The average time for each task is around 2.5 minutes and we pay 0.5 USD per task, hence an hourly rate of $\geq \$12$ per hour. An example of the annotation page is shown in \autoref{fig:human_eval_summ}.

\section{Qualitative Results on Ambiguous Template}\label{sec:results_ambiguous_appendix}
We present two qualitative examples in \autoref{tab:amb_qualitative}. We see that both the original model and \TRIME{} have difficulty in outputting the two correct words as the top two choices. This is also reflected by the cache-only results, where irrelevant words, such as \textit{and}, \textit{I}, \textit{the} get high probabilities. In fact, the cache similarities of the original model are similar to those of \TRIME{}, again indicating that there is no guarantee of well-aligned memories, despite training with the cache. \MODEL{} nevertheless returns the two target words as the top two choices for both the full and cache-only settings, showing that the model benefits from the well-aligned memories through our contrastive objective.

\section{Sample Outputs}

We show sample outputs for prompt continuations in \autoref{fig:sample_prompt}, summarization in \autoref{fig:sample_xsum} and \autoref{fig:sample_cnndm}, and data-to-text in \autoref{fig:sample_data2text}.

\begin{figure*}[ht]
    \centering
    \resizebox{\textwidth}{!}{
    \small
    \begin{tabularx}{\linewidth}{p{\linewidth}}
    \toprule
    \textbf{Prompt}: Every time you die, you are transported to the time of the first event that ultimately caused your death. One day, you are hit by a car and sent 12 years into the past. It's been a strange life; constantly setting \\
    \midrule
    \textbf{GPT2-Large}: up and killing other people. But it's done, and now it ’ s time to settle down in this new world. I ’ d never actually thought about my past life before, I just knew that I went back and lived, but this time there ’ s a complication. I ’ ve been to many places in my life, so why is there such a big problem? It ’ s the weirdest thing, I could go anywhere in time and never get the feeling, but this time it ’ s weird. The cars are just like the last time, the ones that hit me, there ’ s just a hole in the middle of them. I ’ m in a small, abandoned office, and the walls aren ’ t quite made out of cement. I think that ’ s a nice touch ...\\ \\
    \textbf{\TRIME{}}: up and killing other people. But it's done, and now it's time to put the final touches on this latest creation. I'll be frank, I don't really know what I'm supposed to do. A couple months ago, I was on my way to my parents, and there was an accident on the side of the road. It was some kind of accident; the car was hit by a drunk driver, and it fell on my head. Luckily, I was alive. I was given a few months to live, by the hospital. I can live with that, at least until I can figure out what happened. Anyway, I'm now waiting for the day when I meet my next death, and I have to be prepared for the worst. I can try to avoid death ... \\ \\
    \textbf{\MODEL{}}: up and stopping things, trying to prevent every mistake that might happen. If it weren't for the fact that I was always so careful with everything, I might be a normal person who just goes about their business. A normal person like you wouldn't have the ability to go back in time to prevent yourself from ever having died in the first place. Every mistake that might have happened could have killed me then and there, and it wasn't an easy road to travel down. I was lucky enough to have only been born on July 4th, 1996. I've been trying to do the right thing all these years. I've avoided death on more than one  ... \\
    \midrule \midrule
    \textbf{Prompt}: Crown Minotaur Expedition The first week of the expedition was slow going. The terrain was rocky and unforgiving, as if the Mother Nature had forgotten to smooth out this patch of land. The Minotaurs had the most trouble due to the \\
    \midrule
    \textbf{GPT2-Large}: steep slopes of the mountain pass. When a Minotaurel came up the mountain pass, she would pass straight through, just passing through the pass. It was just one of many impassable ridges, that the Miatarel would face to traverse the pass. The only reason it was never impassable was due to the massive, hairy beasts, which covered half of the pass, and caused many people to leave the mountains. This, plus the numerous Minotaurel who passed through the pass would create an increasingly high attrition rate on the expedition. Still, the Miatarel had learned of the pass through the passing ... \\\\
    \textbf{\TRIME{}}: steep slopes of the mountain pass. When a mountain pass was at its highest, the Minotaurs were no longer able to pass into the valley below. It was just too steep. Luckily there was plenty of rock that they could use as a ramp and they would be able to crawl up with their arms, legs, and wings intact. The trip began on the third day after lunch. The trail was still rough, as many of the boulders were nearly a foot high. By now they could barely lift the weight of their bodies, much less their bags full of provisions and equipment. The road ahead of them was now covered in small puddles. This was typical of these ... \\\\
    \textbf{\MODEL{}}: steep slopes of the mountain pass. When a mountain pass was at its highest, the Minotaurs were no longer able to pass into the valley below. It was just too steep. Luckily, they were able to set up camps and set out in the first few days. This was only going to be a matter of time until they realized where they were going. This place was far away from anything they knew of. The only light was the lanterns on their back, and the lanterns were only good for a few minutes. As for how they got there, no one is quite sure. They have not been able to find any of the equipment they carried when they left ... \\
    \bottomrule
    \end{tabularx}
    }
    \caption{Sample outputs for open-ended prompt continuation.}
    \label{fig:sample_prompt}
\end{figure*}

\begin{figure*}[ht]
    \centering
    \resizebox{\textwidth}{!}{
    \small
    \begin{tabularx}{\linewidth}{p{\linewidth}}
    \toprule
    \textbf{Document}: David Lipton, second in command at the IMF, outlined some of these risks in a speech to the National Association for Business Economics in Washington on Tuesday. "The IMF's latest reading of the global economy shows once again a weakening baseline," he said. "We are clearly at a delicate juncture." The comments come after weaker-than-expected trade figures from China showing that exports plunged by a quarter from a year ago. The IMF has already said it is likely it will downgrade its current forecast of 3.4\% for global growth when it next releases its economic predictions in April. The dismal picture is one that has on-going ramifications for businesses and industries that bet on China's growth story. Read more from Karishma: Why a story about bulk shipping matters \\
    \midrule
    \textbf{BART}: The International Monetary Fund (IMF) has warned that the global economy is at a "critical juncture". \\
    \textbf{\TRIME{}}: The International Monetary Fund (IMF) has warned that the global economy is in a "dangerous situation".\\
    \textbf{\MODEL{}}: The International Monetary Fund (IMF) has warned that the global economy is in "a delicate juncture". \\
    \midrule \midrule
    \textbf{Document}: Coventry University's Scarborough campus has been built on the town's former Weaponness Park and Ride site. About 200 students have begun courses at the site, though it is expected to eventually be home to more than 2,000 students. The building, which includes engineering and science labs, a mock law court and a library, is part of a £50m sports and education facility. Professor Craig Gaskell said: "Launching our new state-of-the-art building is a huge milestone for us and demonstrates our commitment to Scarborough and the Yorkshire coast area." A new University Technical College has been built nearby and Scarborough Athletic FC's new 2,000-seater stadium is also under construction on the site. Coventry University also has a campus near London's Liverpool Street Station and recently announced it will open a campus in Dagenham in September 2017. \\
    \midrule
    \textbf{BART}:A university has officially opened its first campus in North Yorkshire.\\
    \textbf{\TRIME{}}: A new university campus has been officially opened in North Yorkshire.\\
    \textbf{\MODEL{}}: A university campus on the Yorkshire coast has opened to the public.\\
    \bottomrule
    \end{tabularx}
    }
    \caption{Sample outputs for XSum summarization.}
    \label{fig:sample_xsum}
\end{figure*}

\begin{figure*}[ht]
    \centering
    \resizebox{\textwidth}{!}{
    \small
    \begin{tabularx}{\linewidth}{p{\linewidth}}
    \toprule
    \textbf{Document}: Bayern Munich will be without versatile left-back David Alaba for the closing stages of the season after he was ruled out for around seven weeks with a torn knee ligament. Alaba, out for three months earlier this season after a partial ligament tear in his right knee, was injured while playing for Austria in their 1-1 draw against Bosnia on Tuesday. 'I am bitterly disappointed,' the 22-year-old told the club's website. 'But I have a big aim and that is to be there at the season finale.' David Alaba is tackled by Emir Spahic's tackle during Austria's 1-1 draw against Bosnia-Herzegovina . Marc Janko (left) celebrates with Martin Harnik and Alaba after he scored against Liechtenstein last Friday . He is set to start rehabilitation work in about a month, Bayern said. Alaba had been in superb form for Bayern since his comeback, helping them stay firmly on track for a treble of titles with some spectacular goals. Bayern, who are top of the Bundesliga with eight games left and through to the German Cup last eight, face Porto in the Champions League quarter-finals later this month. Alaba (left) has been in excellent form for Bayern this season as the Bavarian giants chase a treble of titles . \\
    \midrule
    \textbf{BART}: David Alaba has been ruled out for around seven weeks with a torn knee ligament. Alaba was injured while playing for Austria in their 1-1 draw against Bosnia on Tuesday. Bayern Munich face Porto in the Champions League quarter-finals later this month. \\
    \textbf{\TRIME{}}:David Alaba has been ruled out for around seven weeks with a torn knee ligament. Alaba was injured while playing for Austria in their 1-1 draw against Bosnia on Tuesday. Bayern Munich face Porto in the Champions League quarter-finals later this month. The 22-year-old is set to start rehabilitation work in about a month.\\
    \textbf{\MODEL{}}: David Alaba has been ruled out for seven weeks with a torn knee ligament. Austria defender was injured in their 1-1 draw against Bosnia-Herzegovina. Bayern Munich face Porto in the Champions League quarter-finals later this month. \\
    \midrule \midrule
    \textbf{Document}: Derry Mathews has won the interim WBA lightweight title after being handed a unanimous points decision over Tony Luis in Liverpool. Mathews saw Richar Abril twice pull out of a fight due to illness to be striped of the belt and just this week replacement Ismael Barroso was forced to pull out after failing to obtain a visa. Derry Mathews celebrates after victory over Tony Luis at Echo Arena on Saturday night . Mathews celebrates with his family after claiming the interim WBA lightweight title . Luis was drafted in at the last minute and while the Canadian will have won a lot of fans with his relentless style, he did not win over the judges and Mathews was handed a 114-112, 114-112, 115-112 decision. Mathews made the brighter start and kept Luis contained behind the jab while landing a decent uppercut at the end of the second round. Mathews throws a right punch during interim WBA lightweight title fight in Liverpool . Luis (right) was drafted in after Richar Abril and Ismael Barroso both pulled out . The Liverpudlian continued to still largely control the fight but Luis showed plenty of aggression himself, coming forward and making life difficult for Mathews, landing a good uppercut of his own at the end of the sixth. Luis' pressure and all-action style was beginning to tell on Mathews, who was more and more often going to the ropes and both men were deducted a point in the eighth, Mathews for a low blow and Luis for use of the head. Luis was in the ascendency in the closing rounds and looked to be pulling some points back on the cards but it was not enough as Mathews was handed a tight decision after a thrilling encounter.\\
    \midrule
    \textbf{BART}: Derry Mathews beat Tony Luis in Liverpool on Saturday night. Mathews saw Richar Abril twice pull out of a fight due to illness to be striped of the belt. Ismael Barroso was forced to pull out after failing to obtain a visa.\\
    \textbf{\TRIME{}}: Derry Mathews beat Tony Luis in Liverpool on Saturday night. Mathews saw Richar Abril twice pull out of a fight due to illness. Ismael Barroso was forced to pull out after failing to obtain a visa. Luis was drafted in at the last minute but did not win over the judges.\\
    \textbf{\MODEL{}}: Derry Mathews beat Tony Luis by unanimous points decision at Echo Arena. Mathews saw Richar Abril and Ismael Barroso pull out of the fight due to illness and illness. The Canadian was handed a 114-112,114-112 and 115-112 decision by the judges. Liverpudlian Luis was drafted in at the last minute after the pair pulled out. \\
    \bottomrule
    \end{tabularx}
    }
    \caption{Sample outputs for CNN/DM summarization.}
    \label{fig:sample_cnndm}
\end{figure*}

\begin{figure*}[ht]
    \centering
    \resizebox{\textwidth}{!}{
    \small
    \begin{tabularx}{\linewidth}{p{\linewidth}}
    \toprule
    Table for "1928 vfl season":
    \begin{tabular}{c c c}
    \toprule
    home team score & away team & away team score \\
    \midrule
    11.12 (78) & st kilda & 21.11 (137)\\
    8.9 (57)&geelong&8.8 (56) \\
    11.15 (81)&richmond&10.13 (73) \\
    22.17 (149)&hawthorn&11.13 (79) \\
    18.18 (126)&fitzroy&11.13 (79) \\
    11.17 (83)&essendon&18.11 (119)\\
    \bottomrule
    \end{tabular} \\
    \textbf{Reference}: St Kilda had the highest Score as an Away Team in the 1928 Vfl Season \\
    \midrule
    \textbf{BART}: Hawthorn had the lowest Away Team Score of any team in the 1928 Vfl Season\\
    \textbf{\TRIME{}}: Geelong had the lowest Score of 8.8 (56) while Hawthorn had the highest Score of 11.13 (79) \\
    \textbf{\MODEL{}}: St Kilda was the Away Team with the highest Score in the 1928 Vfl Season \\
    \midrule
    \midrule
    Table for "1893 english cricket season":
    \begin{tabular}{c c}
    \toprule
    name & matches\\
    \midrule
    john hearne & 29 \\
    tom richardson & 23 \\
    johny briggs & 28 \\
    arthur mold & 28 \\
    bill lockwood & 27 \\
    \bottomrule
    \end{tabular} \\
    \textbf{Reference}: John Hearne, played in more Match than any other Player, with 20 9 \\
    \midrule
    \textbf{BART}: Bill Lockwood and Arthur Mold had the same number of Match \\
    \textbf{\TRIME{}}: Bill Lockwood and Arthur Mold both played 27 Match in the 1893 English Cricket Season \\
    \textbf{\MODEL{}}: John Hearne had the most Match with 29\\
    \bottomrule
    \end{tabularx}
    }
    \caption{Sample outputs for data-to-text generation.}
    \label{fig:sample_data2text}
\end{figure*}

\end{document}